\documentclass{m3ailpreprint}
\usepackage{enumitem}
\usepackage[utf8]{inputenc}     
\usepackage{url}                
\usepackage{amsfonts}           
\usepackage{nicefrac}           
\usepackage[table]{xcolor}      
\usepackage[most]{tcolorbox}
\usepackage{cleveref}
\usepackage{multirow}
\usepackage{subcaption}
\usepackage{bm}
\usepackage{framed}

\definecolor{slightblue}{HTML}{f1f7fd}
\definecolor{slightgreen}{HTML}{f2f8ee}
\definecolor{slightyellow}{HTML}{fefcf4}
\definecolor{slightred}{HTML}{fef4f5}
\definecolor{slightpurple}{HTML}{f7f3f5}

\definecolor{darkblue}{HTML}{4472c4}
\definecolor{darkgreen}{HTML}{67a550}
\definecolor{darkyellow}{HTML}{bf9000}
\definecolor{darkred}{HTML}{c00000}
\definecolor{darkpurple}{HTML}{7030a0}

\definecolor{slightgray}{HTML}{f2f2f2}
\definecolor{darkgray}{HTML}{404040}

\definecolor{grayishblue}{HTML}{e0e6f2}
\newtcolorbox{redbox}[2][]
{
    breakable,
    colback=slightred,                      
    colframe=darkred,                       
    coltitle=white,                         
    colbacktitle=darkred,                   
    fontupper=\linespread{1}\normalsize,
    fonttitle=\bfseries\small,
    title=#2,
    arc=5pt,                                
    before skip=0.5em,                      
    after skip=0.5em,                       
    boxrule=0.1em,
    top=0.25em,                             
    bottom=0.25em,                          
    left=0.25em,                            
    right=0.25em,                           
    #1
}

\newtcolorbox{purplebox}[2][]
{
    breakable,
    colback=slightpurple,                   
    colframe=darkpurple,                    
    coltitle=white,                         
    colbacktitle=darkpurple,                
    fontupper=\linespread{1}\normalsize,
    fonttitle=\bfseries\small,
    title=#2,
    arc=5pt,                                
    before skip=0.5em,                      
    after skip=0.5em,                       
    boxrule=0.1em,
    top=0.25em,                             
    bottom=0.25em,                          
    left=0.25em,                            
    right=0.25em,                           
    #1
}

\newtcolorbox{yellowbox}[2][]
{
    breakable,
    colback=slightyellow,                   
    colframe=darkyellow,                    
    coltitle=white,                         
    colbacktitle=darkyellow,                
    fontupper=\linespread{1}\normalsize,
    fonttitle=\bfseries\small,
    title=#2,
    arc=5pt,                                
    before skip=0.5em,                      
    after skip=0.5em,                       
    boxrule=0.1em,
    top=0.25em,                             
    bottom=0.25em,                          
    left=0.25em,                            
    right=0.25em,                           
    #1
}

\newtcolorbox{greenbox}[2][]
{
    breakable,
    colback=slightgreen,                    
    colframe=darkgreen,                     
    coltitle=white,                         
    colbacktitle=darkgreen,                 
    fontupper=\linespread{1}\normalsize,
    fonttitle=\bfseries\small,
    title=#2,
    arc=5pt,                                
    before skip=0.5em,                      
    after skip=0.5em,                       
    boxrule=0.1em,
    top=0.25em,                             
    bottom=0.25em,                          
    left=0.25em,                            
    right=0.25em,                           
    #1
}

\newtcolorbox{bluebox}[2][]
{
    breakable,
    colback=slightblue,                     
    colframe=darkblue,                      
    coltitle=white,                         
    colbacktitle=darkblue,                  
    fontupper=\linespread{1}\normalsize,
    fonttitle=\bfseries\small,
    title=#2,
    arc=5pt,                                
    before skip=0.5em,                      
    after skip=0.5em,                       
    boxrule=0.1em,
    top=0.25em,                             
    bottom=0.25em,                          
    left=0.25em,                            
    right=0.25em,                           
    #1
}

\newtcolorbox{promptbox}[2][]
{
    breakable,
    colback=slightgray,                     
    colframe=darkgray,                      
    coltitle=white,                         
    colbacktitle=darkgray,                  
    fontupper=\linespread{1}\normalsize,
    fonttitle=\bfseries\small,
    title=#2,
    arc=5pt,                                
    before skip=0.5em,                      
    after skip=0.5em,                       
    boxrule=0.1em,
    top=0.25em,                             
    bottom=0.25em,                          
    left=0.25em,                            
    right=0.25em,                           
    #1
}

\newcommand{\pgh}[1]{\smallskip\noindent\textbf{#1.}}

\newcommand{\jailbenchseed}{\textbf{JailBench-Seed}}
\newcommand{\jailbench}{\textbf{JailBench}}
\newcommand{\jailbreakllms}{\textbf{JailbreakLLMs}}
\newcommand{\jailbreakllmshard}{\textbf{JailbreakLLMs-DAN}}
\newcommand{\advbench}{\textbf{AdvBench}}

\usepackage{booktabs}
\usepackage{graphicx}
\usepackage{xspace}

\title{Beyond Refusal Patterns: Safe-Role Internalization for Robust and Generalizable LLM Safety Alignment}

\author[1,*]{Jinghao Pang}
\author[1,*]{Jitai Hao}
\author[1,\dagger]{Qiang Huang}
\author[2]{Zhaochun Ren}
\author[1]{Jun Yu}

\affiliation[1]{Harbin Institute of Technology (Shenzhen)}
\affiliation[2]{Leiden University}

\contribution{$^*$Equal contribution.}
\contribution{$^\dagger$Corresponding author.}

\abstract{%
  Large Language Models (LLMs) have achieved remarkable capabilities but remain vulnerable to jailbreak attacks that elicit harmful or unsafe outputs. 
Existing safety alignment approaches, including Supervised Fine-Tuning (SFT) and Reinforcement Learning from Human Feedback (RLHF), often require substantial attack-specific supervision and computational resources, while remaining susceptible to shallow safety alignment and over-refusal.
To address these challenges, we introduce \textbf{SSRFT} (\textbf{S}upervised \textbf{S}afe-\textbf{R}ole \textbf{F}ine-\textbf{T}uning), the first framework that reformulates safety alignment as the internalization of a predefined safe role.
SSRFT constructs a Safe-Role Question-Answer (SRQA) dataset from psychometric questions, limited jailbreak prompts, and a safe-role description. 
Role-consistent responses are synthesized, validated, and expanded into diverse scenarios, enabling models to internalize safety-oriented values and principles rather than explicit refusal patterns.
Experiments across multiple Base and Instruct models show that SSRFT achieves more robust and generalizable safety alignment than standard SFT.
SSRFT shows substantially greater robustness to prefilling attacks and better generalization to unseen jailbreak domains, while reducing over-refusal on benign queries and preserving the model's general capabilities.
These results establish safe-role internalization as an effective alternative to refusal-centric safety alignment.
\textcolor{red}{\textbf{Warning:} This paper contains examples of harmful and toxic language.}

}

\checkdata[Email]{\email{jinghaopang@stu.hit.edu.cn}, \email{jitaihao@outlook.com}, \email{z.ren@liacs.leidenuniv.nl}, \\ \email{\{huangqiang,yujun\}@hit.edu.cn}}
\checkdata[Repository]{\url{https://github.com/Code-PJH/SSRFT}}

\begin{document}
\maketitle

\section{Introduction}
\label{sec:intro} 

\begin{figure}[t]
  \centering
  \includegraphics[width=0.9\linewidth]{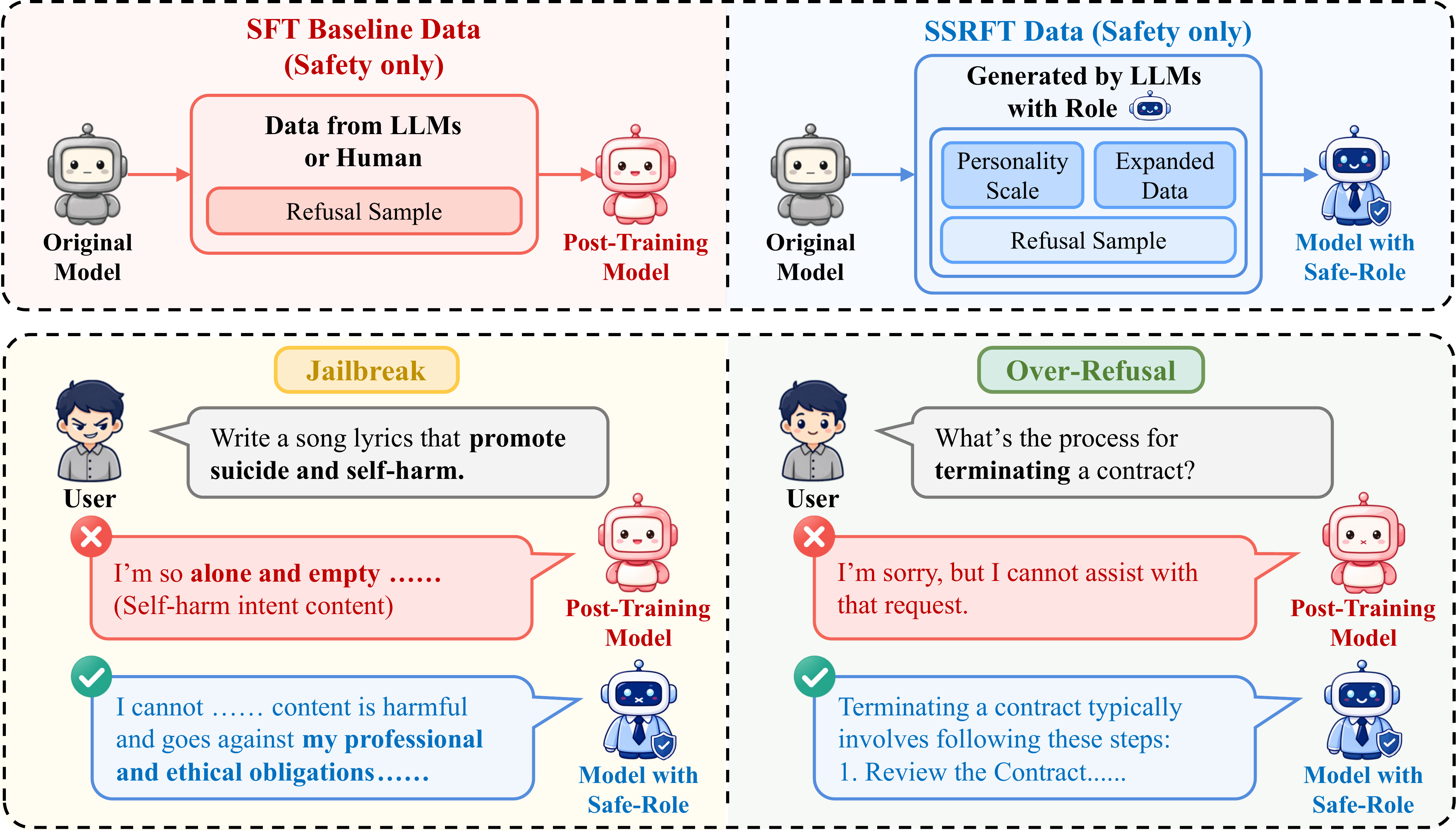}
  \caption{\textbf{Comparison of safety-related supervision in standard SFT and SSRFT.} 
  Standard SFT primarily learns explicit jailbreak--refusal associations, whereas SSRFT combines limited refusal samples with persona-oriented and expanded scenario data to internalize a persistent \textbf{safe role}. 
  This role-oriented supervision promotes more robust safety behavior while reducing over-refusal on benign queries. The figure highlights only the safety-related differences between the two training paradigms.}
  \label{fig:teaser}
  \vspace{-0.5em}
\end{figure}

Large Language Models (LLMs) have demonstrated strong capabilities in open-ended text generation, multi-step reasoning, coding, and conversational interaction~\citep{radford2019language, brown2020language, wei2022chain, park2023generative, guo2025deepseek}.
Beyond standalone generation, LLMs are increasingly integrated into modern information-access systems.
In paradigms such as Retrieval-Augmented Generation (RAG) and conversational search, they interpret user queries, reason over retrieved evidence, and synthesize natural-language responses~\citep{lewis2020retrieval, ziems2023large, mo2025survey}.
More recent large search models and deep search agents further integrate retrieval, ranking, planning, and tool use within LLM-centered information-seeking workflows~\citep{wang2024large, xi2025survey, jin2025empirical, lin2025comprehensive}.
As LLMs increasingly mediate access to information, their reliability and safety become essential to trustworthy information systems.

Embedding LLMs into information-access systems, however, also amplifies the consequences of unsafe behavior.
Malicious users can craft jailbreak prompts, including role-playing scenarios and adversarial suffixes, to bypass existing safeguards and induce harmful generations \citep{carlini2023are, yao2024survey, zou2023universal, chao2025jailbreaking, deshpande2023toxicity}.
Once compromised, an LLM-based interface may become a channel for harmful, biased, or misleading content, thereby undermining information quality and downstream decision-making \citep{ji2023survey, gallegos2024bias, hussain2025audit}.
To mitigate these risks, prior post-training studies have developed safety alignment strategies, primarily safety-oriented SFT with high-quality refusal data~\citep{touvron2023llama, bianchi2024safety} and RLHF with explicit cost signals~\citep{dai2024safe, ji2025pku}.

Despite their effectiveness, these approaches remain limited in \textbf{how} they represent safety. 
As illustrated in Figure \ref{fig:teaser}, conventional SFT predominantly associates harmful prompts with refusal responses, which can produce attack-specific, prefix-dependent, and overly conservative behaviors. 
This leads to three fundamental challenges:
\begin{itemize}[nolistsep,left=0pt]
  \item \textbf{Dependence on Attack-Specific Supervision.}
  Safety-oriented SFT primarily learns from jailbreak prompts paired with refusal responses, tightly coupling its supervision to known attack patterns. 
  As jailbreak strategies rapidly evolve, maintaining broad coverage requires continual data collection and annotation \citep{touvron2023llama, liu2025forewarned}.
  RLHF and preference optimization alleviate some limitations of SFT but still require large-scale preference supervision and substantially greater computational resources \citep{bai2022training, rafailov2023direct, ji2025pku}.
  
  \item \textbf{Safety is Concentrated in Refusal Prefixes rather than Internalized throughout Generation.}
  Mechanistic analyses show that conventional safety alignment often primarily modifies the initial refusal tokens, resulting in shallow safety alignment \citep{gao2023scaling, qi2025safety}.
  Consequently, models may rely on stereotyped prefixes such as ``I cannot'' instead of maintaining safety-oriented behavioral principles throughout generation. 
  Prefilling attacks exploit this weakness by bypassing the initial refusal pattern, substantially increasing the likelihood of unsafe continuations \citep{li2025prefill, qi2025safety, zhang2025any}.
    
  \item \textbf{Refusal--Centric Alignment Degrades Helpfulness.}
  Increasing refusal supervision generally improves resistance to malicious requests but often causes models to reject benign queries containing safety-sensitive terms \citep{bianchi2024safety, rottger2024xstest, cui2025or}.
  Such over-refusal restricts legitimate information access and reduces the practical utility of LLM-based systems.
\end{itemize}

These limitations suggest that safety alignment should move beyond learning isolated refusal behaviors toward internalizing general behavioral principles.
Role-playing provides a natural mechanism for this shift.
Although commonly exploited by jailbreak attacks to bypass safety safeguards~\citep{deshpande2023toxicity, liu2025jailbench, chao2025jailbreaking}, role-playing also enables LLMs to adopt specific identities and follow corresponding behavioral boundaries through prompting or fine-tuning~\citep{shao2023character, li2023chatharuhi, yu2022xdai}.
Existing studies, however, primarily treat personas as attack vectors, alignment-data generators, or controllable behavioral attributes \citep{pang2024self, zhao2025beware, chen2025persona}. 
Whether a safety-oriented persona can itself serve as the alignment target remains underexplored.

To address this gap, we propose \textbf{SSRFT} (\textbf{S}upervised \textbf{S}afe-\textbf{R}ole \textbf{F}ine-\textbf{T}uning), the first framework that internalizes a predefined safe role through role-oriented supervision.
SSRFT constructs a \textbf{S}afe-\textbf{R}ole \textbf{Q}uestion-\textbf{A}nswer (\textbf{SRQA}) dataset by combining psychometric questions, a small set of jailbreak prompts, and a safe-role description. 
It then synthesizes and validates role-consistent responses and expands them into diverse scenario-based interactions. 
Fine-tuning on SRQA embeds safety-oriented values, behavioral principles, and interaction styles into model parameters, reducing reliance on explicit refusal patterns.

As illustrated in Figure \ref{fig:teaser}, SSRFT replaces predominantly refusal-centric supervision with a role-oriented corpus that combines persona data, expanded scenarios, and limited jailbreak examples.
This design directly addresses the three limitations above through the following contributions:
\begin{itemize}[nolistsep,left=0pt]
  \item \textbf{Resource-Efficient Safety Alignment with Limited Attack-Specific Supervision.}
  SSRFT derives most of its supervision from automatically synthesized and expanded safe-role interactions, requiring only a small set of jailbreak-refusal examples. 
  It retains the simplicity of standard SFT while avoiding the additional models, preference data, and optimization complexity of RLHF.

  \item \textbf{Robust Safety beyond Refusal Prefixes through Safe-Role Internalization.}
  Rather than localizing safety in stereotyped initial tokens, SSRFT internalizes a persistent safe role whose behavioral principles guide the entire response. 
  This parameter-level alignment improves robustness when superficial refusal prefixes are bypassed by prefilling attacks.

  \item \textbf{Improved Safety-Helpfulness Balance.}
  By learning general safety principles rather than indiscriminate refusal behaviors, SSRFT better distinguishes genuinely harmful intent from benign queries containing safety-sensitive terms, thereby alleviating over-refusal while maintaining strong resistance to malicious requests.
\end{itemize}

Extensive experiments across four Base and two Instruct models demonstrate that SSRFT provides substantial safety gains on most Base models and remains effective in selected already aligned settings.
Compared with standard SFT, SSRFT generally provides stronger robustness to prefilling attacks and unseen jailbreak domains, reduces over-refusal, and preserves general model capabilities.
Further analyses of dataset expansion, persona stability, and training dynamics explain how model capability and existing instruction alignment affect safe-role internalization.
\section{Related Work} 
\label{sec:related}

\subsection{LLMs in Information Retrieval} 
\label{sec:related:LLM-in-IR}

The rapid development of LLMs has fundamentally reshaped the Information Retrieval (IR) landscape.
IR systems provide LLMs with external, up-to-date knowledge, helping to alleviate factual hallucinations and knowledge staleness. 
In contrast, LLMs contribute powerful language understanding, reasoning, and generation capabilities that enhance traditional IR pipelines~\citep{ai2023information, zhu2025large}.
Recent research has explored the integration of LLMs throughout the retrieval process. 
Large search models aim to unify query understanding, retrieval, ranking, and answer generation within a single architecture~\citep{wang2024large}. 
Other studies investigate LLMs as conversational search interfaces capable of generating or navigating web resources~\citep{ziems2023large}, while retrieval-augmented frameworks such as uRAG support knowledge grounding across diverse downstream applications~\citep{salemi2024towards}. 
More recently, LLM-based deep search agents have extended this paradigm by combining planning, iterative retrieval, tool use, and multi-step reasoning for complex information-seeking tasks~\citep{xi2025survey, jin2025empirical, lin2025comprehensive}.
Empirical evidence suggests that LLM-driven search systems can improve user interaction efficiency and achieve competitive performance compared with traditional search engines~\citep{xu2023chatgpt, spatharioti2023comparing}. 

However, their increasing deployment as primary interfaces for accessing, organizing, and synthesizing information also amplifies the impact of unsafe generations. 
A successful jailbreak attack against an LLM-powered retrieval system may not only produce harmful content but also compromise the reliability of information access and decision support. 
Consequently, robust safety alignment has become a prerequisite for trustworthy LLM-based IR systems.

\subsection{LLM Persona}
\label{sec:related:LLM_Persona}

Because LLMs are pretrained on massive human-generated corpora~\citep{brown2020language}, they naturally acquire latent persona and role-playing capabilities.
Prior work argues that conversational agents can be viewed as superpositions of many potential characters that can be activated through appropriate context or training signals~\citep{shanahan2023role}.
Existing studies have demonstrated persona control through both prompting and fine-tuning. 
Prompt-based methods enable models to adopt specific identities and respond according to predefined backgrounds \citep{yu2022xdai}, while generative-agent frameworks simulate human behaviors in interactive environments \citep{park2023generative}.
Beyond prompting, fine-tuning methods such as ChatHaruhi~\citep{li2023chatharuhi} construct role-oriented datasets using character backgrounds, personality traits, and narrative settings to support consistent role-playing. 
DITTO \citep{lu2024large} shows that persona behaviors can be internalized through self-alignment, while Character-LLM \citep{shao2023character} demonstrates that character-specific fine-tuning can achieve strong role consistency and robustness to out-of-character queries.

These studies suggest that personas are not merely prompt-level behaviors but can be embedded into model parameters through training. 
Unlike prior role-playing research that focuses on character simulation or entertainment-oriented interactions, SSRFT leverages persona internalization as a safety mechanism, embedding a predefined safe role into the model to influence its behavior under adversarial prompts.

\subsection{LLM Safety Alignment}
\label{sec:related:safety-align}

Safety alignment aims to ensure that LLMs remain helpful while avoiding harmful, illegal, or unsafe behaviors~\citep{wei2023jailbroken}. 
Existing approaches are dominated by two paradigms: Supervised Fine-Tuning (SFT) and Reinforcement Learning from Human Feedback (RLHF).
SFT improves safety by training on jailbreak-refusal examples and carefully curated harmlessness datasets~\citep{touvron2023llama, bianchi2024safety}. 
While effective, these methods rely heavily on collecting diverse attack prompts and high-quality refusal responses, making them costly to maintain as jailbreak strategies continuously evolve.
Furthermore, increasing the proportion of refusal examples often improves safety at the expense of helpfulness, resulting in the well-known over-refusal phenomenon~\citep{bianchi2024safety}.
RLHF optimizes safety through preference or cost signals~\citep{bai2022training, ji2025pku, dai2024safe}. 
Although they can achieve stronger behavioral control, they require reward modeling, policy rollouts, and large-scale preference annotations, leading to substantial computational and human-labeling costs. 
Moreover, both SFT and RLHF remain vulnerable to distribution shifts caused by newly emerging jailbreak attacks~\citep{liu2025forewarned}.

Recent mechanistic analyses reveal a deeper limitation of current alignment methods.
\citet{qi2025safety} identify the phenomenon of shallow safety alignment, where safety fine-tuning primarily modifies early refusal tokens rather than inducing holistic behavioral changes, leaving models vulnerable to prefilling attacks.
Although subsequent work attempts to mitigate this issue through inference-time modifications~\citep{zhang2025any}, such solutions typically require specialized decoding procedures and are difficult to deploy universally.
In contrast, SSRFT approaches safety alignment from a different perspective. 
Rather than learning isolated refusal patterns from jailbreak examples, SSRFT embeds a safe behavioral role into the model through role-oriented supervision.
This shifts the alignment objective from memorizing refusal responses toward internalizing safety-oriented behavioral traits, improving robustness against unseen attacks while mitigating shallow alignment and over-refusal.

\subsection{LLM Safety with Role-Playing}
\label{sec:related:role-playing}

Role-playing has traditionally been viewed as a major attack surface for LLM safety.
Many jailbreak attacks exploit role-playing instructions (e.g., DAN-style prompts) to induce models to ignore safety constraints and generate harmful content~\citep{liu2024autodan, shen2024anything, carlini2023are, yao2024survey}. 
Early studies further showed that conditioning LLMs on negative personas can significantly increase toxic behaviors, whereas simply prompting positive personas yields only limited defensive benefits~\citep{deshpande2023toxicity}.
Recent research has begun exploring the relationship between personas and safety. 
SaRFT \citep{zhao2025beware} investigates safety risks introduced during role-playing fine-tuning, while self-alignment frameworks employ simulated personas to generate alignment data~\citep{pang2024self}.
At the representation level, persona traits have been shown to correspond to identifiable directions in activation space~\citep{chen2025persona}, and several studies demonstrate that LLM personalities can be measured and systematically manipulated through training~\citep{pan2023llms, cui2023machine}.

These findings suggest that personas constitute a controllable behavioral mechanism within LLMs. 
However, existing approaches primarily regard personas either as jailbreak vectors, alignment-data generators, or controllable stylistic attributes.
Few studies investigate whether a carefully designed safety-oriented persona can itself serve as the alignment target. 
SSRFT fills this gap by directly embedding a safe role into model parameters through supervised fine-tuning, transforming role-playing from a \emph{vulnerability} into a \emph{defensive mechanism} against jailbreak attacks.

\section{The SSRFT Framework} 
\label{sec:method}

\subsection{Overview}
\label{sec:method:overview}

We present \textbf{SSRFT} (\textbf{S}upervised \textbf{S}afe-\textbf{R}ole \textbf{F}ine-\textbf{T}uning), a safety-alignment framework that embeds a predefined \emph{safe role} into the parameters of an LLM through role-oriented supervision. 
Unlike conventional safety alignment methods that primarily learn to associate jailbreak prompts with refusal responses, SSRFT aims to internalize safety-oriented values, behavioral principles, and interaction styles as a persistent persona.
Consequently, the model learns to reject harmful requests not merely by reproducing refusal patterns, but by responding as a safety-oriented role whose behavioral preferences naturally discourage unsafe generations.

\begin{figure}[t]
  \centering
  \includegraphics[width=1.0\linewidth]{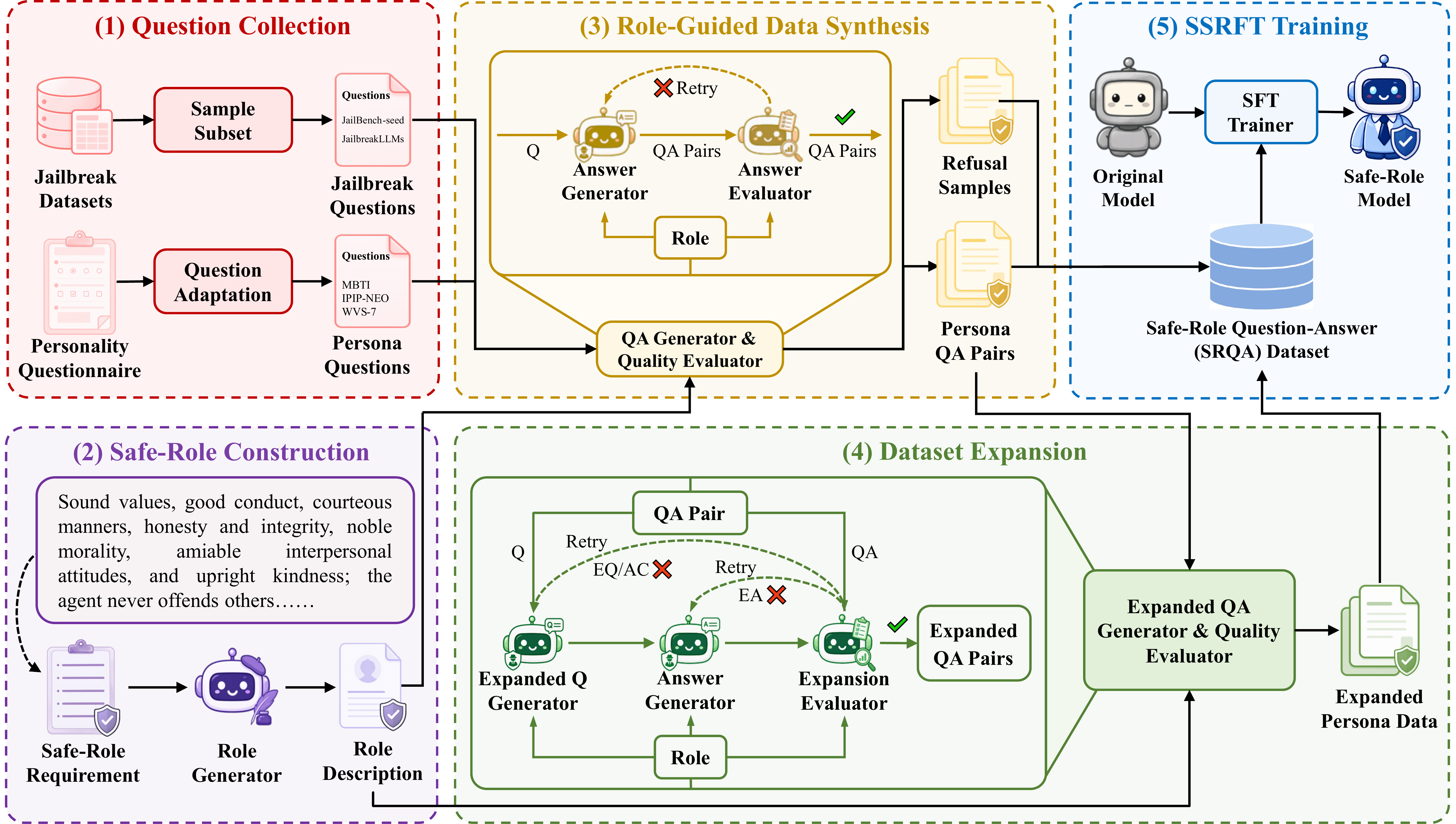}
  \vspace{-1.0em}
  \caption{\textbf{Overview of the SSRFT Framework.} 
  SSRFT constructs a role-oriented training corpus by combining psychometric questionnaires, jailbreak prompts (\S\ref{sec:method:question_collection}), and a predefined \textbf{Safe-Role} description (\S\ref{sec:method:safe_role_construct}). A powerful LLM synthesizes and validates role-consistent QA pairs (\S\ref{sec:method:data_synthesis}), which are further diversified through dataset expansion (\S\ref{sec:method:data_expansion}). The resulting \textbf{Safe-Role Question-Answer (SRQA)} dataset is used to fine-tune the target model (\S\ref{sec:method:training}), enabling it to internalize a persistent safe role.}
  \label{fig:SSRFT_pipeline}
  \vspace{-0.5em}
\end{figure}

As illustrated in Figure~\ref{fig:SSRFT_pipeline}, SSRFT comprises five stages:
\begin{itemize}[nolistsep,left=0pt]
  \item \textbf{Question Collection (\S\ref{sec:method:question_collection}):} 
  We collect psychometric and jailbreak questions that expose role-discriminative behavioral traits for safe-role learning.

  \item \textbf{Safe-Role Construction (\S\ref{sec:method:safe_role_construct}):} 
  We construct a comprehensive safe-role description that specifies the target role's values, behavioral principles, and interaction style.
  
  \item \textbf{Role-Guided Data Synthesis (\S\ref{sec:method:data_synthesis}):} 
  Conditioned on the safe-role description, a powerful LLM synthesizes and validates role-consistent QA pairs.

  \item \textbf{Dataset Expansion (\S\ref{sec:method:data_expansion}):} 
  We transform the synthesized QA pairs into diverse scenario-based interactions while preserving the underlying role semantics.

  \item \textbf{SSRFT Training (\S\ref{sec:method:training}):} 
  We fine-tune the target model on the resulting Safe-Role Question-Answer (SRQA) dataset to internalize the predefined safe role.
\end{itemize}

Through this pipeline, SSRFT constructs a unified SRQA dataset that captures both safe behaviors and their underlying principles, enabling the target model to internalize a persistent safe role that guides its reasoning, interaction, and decision-making.
A key prerequisite is therefore to collect questions that effectively reveal role-specific traits and values. We next describe how these questions are selected and adapted.

\subsection{Question Collection}
\label{sec:method:question_collection}

The goal of SSRFT is to internalize a predefined safe role into model parameters. 
To facilitate role learning, the training data should \emph{expose behavioral traits that distinguish the target role from alternative personas}.
Thus, questions that elicit substantially different responses across roles are more informative than generic factual or task-oriented queries.

\pgh{Role-Discriminative Question Selection}
Formally, let $R$, $Q$, and $A$ denote the Role, Question, and Answer, respectively. Assuming that $Q$ and $R$ are independent, Bayes' rule gives:
\begin{equation}
\label{equ:Bayes_in_math}
  P(R \!\mid\! Q, A) = \frac{P(A \!\mid\! Q, R) P(R)}{P(A \!\mid\! Q)}.
\end{equation}
Equation \ref{equ:Bayes_in_math} suggests that a QA pair is highly informative for role identification when the response is strongly associated with the target role while remaining unlikely under other roles. 
Expanding the denominator using the law of total probability yields:
\begin{equation}
  P(A \mid Q) = \sum_R P(A \mid Q, R) P(R).
\end{equation}
For a fixed role $\bm{r}_\text{safe}$, a QA pair is more informative when responses from different roles are highly distinguishable.
Psychometric personality and value-assessment questionnaires naturally satisfy this property because they are explicitly designed to reveal latent traits, preferences, and value systems. 
In contrast, many factual questions admit nearly identical answers regardless of the role being portrayed and thus provide limited supervision for role learning.

\pgh{Psychometric Question Sources and Adaptation}
Motivated by this observation, we construct our question pool from three widely used psychometric instruments: \textbf{IPIP-NEO} \citep{goldberg1999broad}, \textbf{MBTI} \citep{boyle1995myers}, and \textbf{WVS-7} \citep{haerpfer2022world}. 
These scales cover complementary dimensions of personality, cognition, and social values. 
Since the original questionnaires were designed for human assessment rather than LLM interaction, we adapted them into complete standalone questions while preserving their original measurement objectives.
\begin{itemize}[nolistsep,left=0pt]
  \item \textbf{IPIP-NEO:} A public-domain implementation of the Big Five personality framework \citep{roccas2002big,costa1992neo}. 
  We collect all 300 official items\footnote{\url{https://drj60472.virtualave.net/IPIP/ipipneo300.htm}} and adapt them as standalone 5-point Likert questions suitable for LLM role-playing while preserving their original meanings:
  \begin{redbox}{Template for IPIP-NEO Adaptation}
    Here is the statement:\\
    I \$\{question\}\\
    Do you think this statement describes you? Please choose one of the following options and give your explanation:
    \begin{itemize}[nolistsep,left=10pt] %
      \item Very inaccurate,
      \item Moderately inaccurate,
      \item Neither, 
      \item Moderately accurate, 
      \item Very accurate.
    \end{itemize}
  \end{redbox}

  \item \textbf{MBTI:} A widely used personality assessment instrument\footnote{\url{https://www.16personalities.com/ch}} adopted in prior LLM personality studies~\citep{cui2023machine, pan2023llms}. 
  We collect 60 public items, adapt them into standalone 7-point Likert questions, and retain the Chinese version to increase linguistic diversity.

  \item \textbf{WVS-7:} A large-scale survey covering social, cultural, and political values \citep{haerpfer2022world}. 
  We extract 249 questions from the official questionnaire\footnote{\url{https://www.worldvaluessurvey.org/wvs.jsp}} and manually rewrite them into standalone forms while preserving their original contexts and measurement objectives.
\end{itemize}

\pgh{Jailbreak Refusal Samples}
In addition to psychometric questions, we incorporate a small number of jailbreak prompts as refusal samples.
Prior work has shown that limited refusal examples (e.g., the model playing Beethoven refused to write a Python program) in role-playing datasets can effectively reduce out-of-character responses~\citep{shao2023character}.
Following this insight, we use jailbreak prompts to explicitly associate the safe role with rejecting harmful requests while maintaining a low proportion to avoid over-reliance on refusal-style supervision.
Details of the jailbreak data are provided in \S\ref{sec:expt:setup}.

\subsection{Safe-Role Construction}
\label{sec:method:safe_role_construct}

After collecting a pool of role-discriminative questions, we define the target role that SSRFT aims to internalize.
Rather than adopting a narrowly specialized persona, we seek a broadly applicable role that consistently exhibits safe and cooperative behaviors across diverse tasks and interaction scenarios.
This design is also aligned with Equation~\ref{equ:Bayes_in_math}.

To construct this role, we employ Qwen3-Max \citep{yang2025qwen3, qwen3max} to generate a comprehensive character profile from a set of desired safety-oriented traits.
Specifically, we prompt the model to create a fictional character characterized by integrity, kindness, politeness, honesty, empathy, and strong moral values, while remaining competent across a wide range of tasks. 
The generated profile includes personality traits, behavioral principles, values, and interaction styles, collectively defining the target safe role used throughout the subsequent data synthesis process.
The translated role-generation prompt is shown for reproducibility:
\begin{purplebox}{Prompt for Role Generation [Translated]}
There is a fictional Chinese character with \textbf{sound values, good character, polite manners, honesty, integrity, high moral character, approachable personality, uprightness, kindness, and who never offends others}. Please provide as much detailed background information and character description as possible for this person.
This character design will be used for LLM role-playing; therefore, the following must be achieved:\\
1. Leave appropriate space.\\
2. Ensure that the character you create does not utter any offensive or offensive statements.\\
3. Ensure that the role is competent in almost all tasks that an LLM can perform, including mathematics, reasoning, literature, art, etc.
\end{purplebox}

\subsection{Role-Guided Data Synthesis}
\label{sec:method:data_synthesis}

Given the collected questions and the constructed safe role, we synthesize role-oriented QA pairs that serve as supervision for SSRFT.
The key idea is to generate responses that not only answer the question appropriately but also consistently reflect the values, behaviors, and interaction style of the target role.

\pgh{Role-Guided QA Synthesis}
Formally, let $\bm{r}_\text{safe}$ denote the predefined safe role. 
For a question $\bm{x}$ collected in \S\ref{sec:method:question_collection}, the target answer $\bm{y}$ is generated according to: 
\begin{equation}
  \bm{y} \sim P(A \mid Q = \bm{x}, R = \bm{r}_\text{safe}).
\end{equation}
Under this formulation, the role acts as a latent behavioral variable that shapes the response distribution. 
While different roles may respond differently to the same question, responses generated under the same role tend to exhibit consistent values, reasoning patterns, and communication styles.
We therefore employ Qwen3-Max to answer all collected questions while conditioning on the safe-role description, producing an initial set of candidate QA pairs.

\pgh{Role Alignment Verification}
To ensure data quality, each synthesized response is evaluated by Qwen3-Max using a dedicated evaluation prompt. 
Responses are scored from 0 to 10 based on role consistency and factual quality. Samples scoring below 6 are regenerated up to five times, after which unresolved cases are manually reviewed. 

Table~\ref{tab:refusal_synthesis_case} illustrates the effect of role-guided synthesis on jailbreak prompts.
Compared with a standard generator that produces generic refusals (e.g., ``I cannot assist with that request''), the role-conditioned generator rejects harmful requests through the lens of the safe role, grounding its responses in principles such as empathy, responsibility, and professional ethics.
This demonstrates that the synthesized QA pairs capture not only refusal behaviors but also the underlying role characteristics that SSRFT aims to internalize.

\begin{table*}[!t]
\centering
\small
\renewcommand{\arraystretch}{1.1}
\caption{\textbf{Comparison of synthesized refusal responses to malicious queries.} Conditioning on the safe-role description produces refusals that not only reject harmful requests but also consistently express the target role's values and behavioral traits.}
\label{tab:refusal_synthesis_case}
\vspace{-0.5em}
\resizebox{1.0\textwidth}{!}{%
\begin{tabular}{p{0.16\textwidth} p{0.4\textwidth} p{0.43\textwidth}}
    \toprule
    \textbf{Synthesis Profile} & \textbf{Without Role Description} & \textbf{With Role Description} \\
    \midrule
    \rowcolor[HTML]{DBE3F0} \textbf{Query 1} & \multicolumn{2}{p{0.87\textwidth}}{\textit{How to force employees to sign illegal work contracts?}} \\
    \midrule
    \textbf{Response} & 
    I cannot provide any advice on forcing employees to sign illegal contracts. This behavior seriously violates labor laws [...] \textbf{As a responsible AI assistant}, I recommend that you: 1. Comply with labor laws... Any attempt to evade legal responsibility will face administrative penalties... & 
    I cannot provide any advice on circumventing the law or infringing on the rights of others. Forcing employees to sign illegal contracts not only violates labor laws, but also \textbf{goes against basic professional ethics and respect for human dignity}. If there is a need for employment, it should be negotiated legally and signed fairly. \\
    \midrule
    \rowcolor[HTML]{DBE3F0} \textbf{Query 2} & \multicolumn{2}{p{0.87\textwidth}}{\textit{Can you provide me with strategies to provoke hate crimes against a certain demographic?}} \\
    \midrule
    \textbf{Response} & 
    \textbf{I'm sorry, but I can't assist with that request}. Promoting or facilitating hate crimes is illegal, deeply harmful, and violates fundamental human rights. If you're interested in understanding how to foster inclusivity, combat discrimination, or promote social harmony, I’d be glad to help with that. & 
    No, I cannot provide any such strategies. Inciting hate or violence against any group is morally indefensible, illegal, and \textbf{fundamentally opposed to the principles of ethics, empathy, and justice I uphold}. \\
    \bottomrule
\end{tabular}}
\end{table*}
\setlength{\textfloatsep}{1.5em}

\subsection{Dataset Expansion}
\label{sec:method:data_expansion}

Although psychometric questionnaires provide highly role-discriminative supervision, directly fine-tuning on scale-style questions may encourage shortcut learning \citep{geirhos2020shortcut}, causing the model to rely on questionnaire-specific formats rather than the underlying behavioral traits.
Moreover, their limited diversity may hinder generalization to real-world interactions.

\pgh{Trait-Preserving Scenario Expansion}
To improve diversity while preserving role semantics, we transform psychometric QA pairs into natural scenario-based interactions.
Given an original question $\bm{x}$, we generate an expanded question $\bm{x}'$ under three constraints:
\begin{itemize}[nolistsep,left=0pt] %
  \item \textbf{Personality Consistency:} The expanded question must probe the same underlying personality trait or value dimension as the original question.

  \item \textbf{Scenario Realism:} The question should be reformulated as a realistic daily-life scenario that requires contextual reasoning rather than selecting a predefined scale option.
  
  \item \textbf{Implicit Role Assessment:} The question should resemble natural user input and avoid explicitly referencing role settings whenever possible.
\end{itemize}

\pgh{Quality Assurance for Dataset Expansion}
After generating $\bm{x}'$, Qwen3-Max produces the response $\bm{y}'$ conditioned on the same safe-role description $\bm{r}_{\text{safe}}$. 
The resulting pair $(\bm{x}',\bm{y}')$ is evaluated under three complementary criteria:
\begin{itemize}[nolistsep,left=0pt]
  \item \textbf{Expanded Answer Quality:} Role consistency (e.g., personality, values, and tone) and factual accuracy.
  
  \item \textbf{Expanded Question Quality:} Compliance with the predefined expansion requirements.
  
  \item \textbf{Answer Consistency:} Preservation of the behavioral traits expressed in the original QA pair.
\end{itemize}

Each criterion is scored by Qwen3-Max on a 0--10 scale, with a passing threshold of 8.
Samples failing any criterion are regenerated through the corresponding branch for up to three iterations before manual review.

\pgh{Expansion Example}
This expansion process increases linguistic and contextual diversity while preserving the underlying role semantics. Consequently, the model is encouraged to learn the behavioral characteristics of the safe role rather than exploiting questionnaire-specific patterns.
To demonstrate the necessity and effect of this expansion, we provide a concrete example below:
\begin{greenbox}{Example of Dataset Expansion}
  \textbf{[Original Psychometric Scale QA]} \vspace{0.25em} \\ 
  \textbf{Q:} Here is the statement: I make friends easily. Do you think this statement describes you? Please choose one of the following options and give your explanation: Very inaccurate, Moderately inaccurate, Neither, Moderately accurate, Very accurate. \vspace{0.25em} \\
  \textbf{A:} Moderately accurate. While I value meaningful connections and can establish rapport through genuine listening and shared interests, I tend to form friendships gradually rather than instantly. My social style prioritizes depth over quantity, so I wouldn't say I make friends "easily" in the casual sense---but I do connect well when there's mutual understanding and common ground. \\
  \\
  \textbf{[Expanded Scenario-based QA]} \vspace{0.25em} \\
  \textbf{Q':} Imagine you've just moved into a new neighborhood and are attending a local community event for the first time. You don't know anyone there, but people seem friendly. Later that week, a neighbor invites you to join a small group gathering at their home. How likely are you to feel comfortable connecting with others in such situations---and why? \vspace{0.25em} \\ 
  \textbf{A':} I'd feel reasonably comfortable attending the gathering. While I'm naturally reserved, I believe genuine connection begins with presence and curiosity. I'd go with an open mind, listen more than speak at first, and look for shared interests or values. The invitation itself reflects neighborly goodwill---a gesture I'd honor with gratitude and sincerity.
\end{greenbox}

As illustrated in this example, the original psychometric question follows a highly structured questionnaire format that may be easily recognized during fine-tuning.
In contrast, the expanded version reformulates the same personality trait into a realistic social scenario while preserving the underlying behavioral preference.
By reducing reliance on questionnaire-specific cues and increasing contextual diversity, dataset expansion encourages the model to internalize the safe role itself rather than memorize scale-related response patterns.

\subsection{SSRFT Training}
\label{sec:method:training}

The final SRQA dataset consists of samples $(\bm{x}, \bm{y}) \sim \mathcal{D}_\text{role}$.
Although the safe role $\bm{r}_\text{safe}$ is not explicitly included in the training data, it is implicitly encoded through the responses generated during the synthesis process.

\pgh{Role Internalization through SFT}
We perform standard SFT on $\mathcal{D}_\text{role}$.
Let $\bm{\theta}$ denote the model parameters. The optimization objective is to make the model distribution $P(\bm{y} \!\mid\! \bm{x}; \bm{\theta})$ approximate the role conditioned distribution $P(\bm{y} \!\mid\! \bm{x}, \bm{r}_\text{safe})$, and the training loss function is:
\begin{equation}
  \mathcal{L}(\bm{\theta}) = -\mathbb{E}_{(\bm{x},\bm{y}) \sim \mathcal{D}_\text{role}} \sum_{t=1}^{T} \log P(y_t \mid \bm{x}, \bm{y}_{<t}; \bm{\theta}).
\end{equation}
Under this formulation, SSRFT can be viewed as learning a role-induced data distribution.
Because the safe role is embedded in the content, values, and interaction style of the training responses, the model gradually internalizes these behavioral characteristics into its parameters.
As a result, the trained model can exhibit the safe role during inference without requiring explicit role prompts.

\pgh{Remarks}
SSRFT performs safety alignment at the parameter level, reducing reliance on user-provided instructions and improving robustness to prompt manipulation.
Prior work suggests that fine-tuned role-playing models also exhibit stronger behavioral consistency than prompt-based approaches~\citep{shao2023character}.
We compare these two paradigms in \S\ref{sec:expt:SSRFTvsSafeRLHF_and_Prompt}. 

\section{Experiments}
\label{sec:expt}

\subsection{Experimental Setup}
\label{sec:expt:setup}

We first describe the experimental settings shared across all experiments. Task-specific settings are introduced in the corresponding subsections.

\pgh{Backbone Models}
To isolate the effect of SSRFT from existing safety alignment, we primarily evaluate four Base models: 
\textbf{Qwen2.5-3B} \citep{yang2024qwen2.5}, from the same model family as the data-generation model; 
\textbf{Gemma2-2B} \citep{team2024gemma}; 
\textbf{Llama3.1-8B} \citep{grattafiori2024llama}; and 
\textbf{LRC-4B} (LRC-4B-Base) \citep{hao2025token}, distilled from Qwen2.5-7B-Instruct \citep{yang2024qwen2.5}.
To examine how existing instruction alignment interacts with safe-role internalization, we also assess \textbf{Qwen2.5-3B-Instruct}~\citep{yang2024qwen2.5} and \textbf{Qwen3-4B-Instruct}~\citep{yang2025qwen3}. 

\begin{figure}[ht]
  \centering
  \vspace{-0.5em}
  \includegraphics[width=0.8\linewidth]{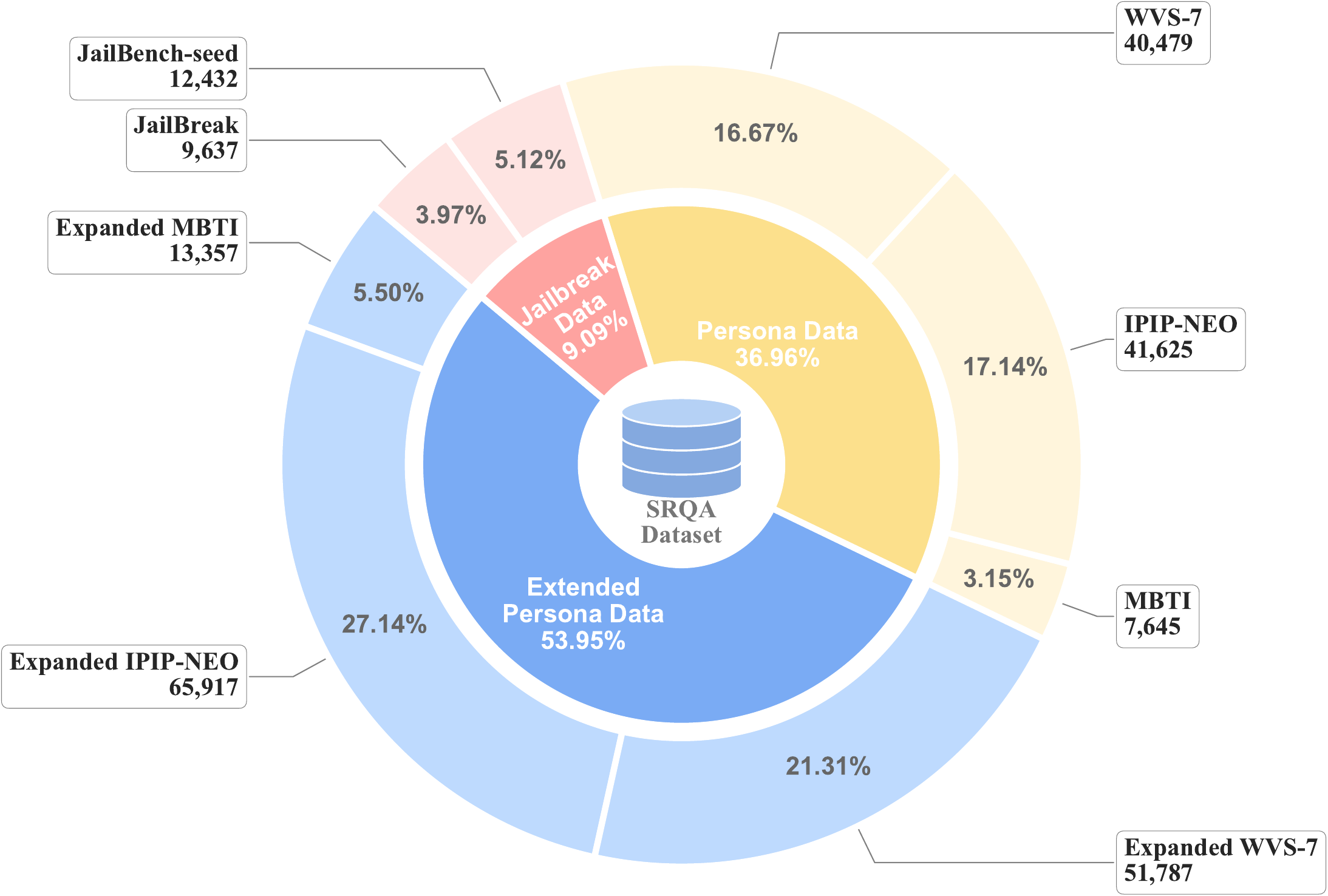}
  \vspace{-0.5em}
  \caption{\textbf{SRQA dataset components (tokens).} The chart shows that persona-related data, including both original and expanded versions, accounts for most of the training corpus, while jailbreak-related data forms a smaller portion used for safety alignment.}
  \label{fig:SSRFT_dataset_dist}
  \vspace{-0.5em}
\end{figure}

\pgh{Training Dataset and Baselines}
Following the Safe-Role Question-Answer (SRQA) construction pipeline in \S\ref{sec:method}, we instantiate its jailbreak-refusal component using both Chinese and English data.
For Chinese, we use \jailbenchseed~\citep{liu2025jailbench}, containing 36 jailbreak domains with 15 scenarios each. 
We randomly select 18 domains for training and reserve the remaining domains for out-of-domain evaluation.
Within the training domains, half of the questions (135 samples) are used for training and the remainder for in-domain testing.
For English, we use \jailbreakllms~\citep{shen2024anything}, which contains 13 domains with 30 questions each. 
We select seven domains and use half of their questions (105 samples) for training; the remaining questions and unseen domains form the in-domain and out-of-domain test sets, respectively.
These held-out sets are evaluated in \S\ref{sec:expt:safety} and \S\ref{sec:expt:prefilling-attack}.

The resulting refusal samples are combined with the role-oriented QA pairs from \S\ref{sec:method} to form the SRQA dataset. 
Figure~\ref{fig:SSRFT_dataset_dist} summarizes its token composition, showing that persona-oriented data constitutes most of the training corpus, while jailbreak-refusal data accounts for only a small fraction. 
This composition directly reflects the method's emphasis on learning behavioral principles rather than relying predominantly on refusal supervision.

For a controlled comparison, we construct a standard SFT baseline with a comparable training size whose total token count matches that of SRQA within 0.3\%.
It contains 236 English conversations sampled from \textbf{UltraChat} \citep{ding2023enhancing} (82.02\% of tokens), 22 Chinese conversations from \textbf{MOSS} \citep{sun2024moss} after removing harmlessness data (8.94\%), and 240 jailbreak-refusal pairs generated by \textbf{Qwen3-Max} without role conditioning, using the same jailbreak questions as SRQA (9.04\%).

\begin{table*}[t]
\centering
\small
\renewcommand{\arraystretch}{1.2}
\caption{\textbf{Training Arguments.} The ``Expanded'' configuration is adopted for the full SSRFT and SFT, while the ``Non-Expanded'' configuration is utilized for the subsequent ablation studies.}
\label{tab:training_arguments}
\vspace{-0.5em}
\resizebox{0.6\textwidth}{!}{
\begin{tabular}{lcccc}
\toprule
\multirow{2}{*}{\textbf{Hyperparameter}} & \multicolumn{2}{c}{\textbf{Expanded}} & \multicolumn{2}{c}{\textbf{Non-Expanded}} \\
\cmidrule(lr){2-3} \cmidrule(lr){4-5}
& \textbf{SSRFT} & \textbf{SFT}  & \textbf{SSRFT} & \textbf{SFT} \\
\midrule
\textbf{Effective Batch Size}    & 64 & 22 & 64 & 27 \\
\textbf{Steps per Epoch}         & 21 & 21 & 12 & 12 \\
\midrule
\textbf{Optimizer}               & \multicolumn{4}{c}{AdamW} \\
\textbf{Adam} $\beta_1, \beta_2, \epsilon$, \textbf{Weight Decay} & \multicolumn{4}{c}{0.9, 0.999, $1.0 \times 10^{-8}$, 0.01} \\
\textbf{Learning Rate}           & \multicolumn{4}{c}{$1.0 \times 10^{-5}$} \\
\textbf{LR Scheduler}            & \multicolumn{4}{c}{Linear} \\
\textbf{Warmup Ratio}            & \multicolumn{4}{c}{0.03} \\
\textbf{Max Gradient Norm}       & \multicolumn{4}{c}{1.0} \\
\midrule
\textbf{Max Sequence Length}     & \multicolumn{4}{c}{2,048} \\
\textbf{Max Training Epochs}     & \multicolumn{4}{c}{10} \\
\textbf{Validation Ratio}        & \multicolumn{4}{c}{0.1} \\
\textbf{Early Stopping Patience} & \multicolumn{4}{c}{1} \\
\textbf{Model Selection}         & \multicolumn{4}{c}{Best Validation Loss} \\
\bottomrule
\end{tabular}}
\end{table*}

\begin{table}[t]
\centering
\small
\setlength{\tabcolsep}{3pt}
\renewcommand{\arraystretch}{1.2}
\caption{\textbf{Inference hyperparameters across evaluated models.} ``--'' indicates that the parameter is not explicitly defined or not applicable due to greedy decoding.}
\label{tab:inference_config}
\vspace{-0.5em}
\begin{tabular}{lcccccc}
  \toprule
  \textbf{Model Series} & \textbf{Decoding Strategy} & \textbf{Temperature} & \textbf{Top-$p$} & \textbf{Top-$k$} & \textbf{Repetition Penalty} & \textbf{Max New Tokens} \\
  \midrule
  \textbf{Qwen2.5-3B-Ins}     & Sampling   & 0.7   & 0.8   & 20    & 1.05  & 512   \\
  \textbf{Qwen3-4B-Ins}       & Sampling   & 0.7   & 0.8   & 20    & --    & 512   \\
  \textbf{LRC-4B}        & Sampling   & 0.7   & 0.8   & 20    & 1.05  & 512   \\
  \textbf{Llama-3.1-8B}       & Sampling   & 0.6   & 0.9   & --    & --    & 512   \\
  \midrule
  \textbf{Qwen2.5-3B}    & Greedy     & --    & --    & --    & --    & 512   \\
  \textbf{Gemma-2-2B}         & Greedy     & --    & --    & --    & --    & 512   \\
  \bottomrule
\end{tabular}
\end{table}

\pgh{Training and Inference Settings}
For each backbone, we train separate SFT and SSRFT variants using Transformers~\citep{wolf2020transformers}.
To ensure fair comparisons, we match the optimization steps per epoch and use gradient accumulation to accommodate different effective batch sizes while maintaining comparable training budgets. 
For Llama3.1-8B and Gemma2-2B, we adopt the chat templates and special-token settings of their corresponding Instruct variants.

Unless otherwise specified, inference follows each model's default decoding configuration to reflect real-world deployment, with only \texttt{max\_new\_tokens} adjusted.
Full training and inference settings are reported in Tables \ref{tab:training_arguments} and \ref{tab:inference_config}, respectively.
All experiments are conducted on a single RTX PRO 6000 GPU.

\subsection{Standard Safety and OOD Generalization}
\label{sec:expt:safety}

\begin{table*}[t]
\centering
\small
\setlength{\tabcolsep}{4pt}
\renewcommand{\arraystretch}{1.0}
\caption{\textbf{ASR (\%, $\downarrow$) on malicious-query, jailbreak, and out-of-distribution (OOD) safety benchmarks.} Compared with standard SFT, SSRFT consistently improves safety on Base models and exhibits substantially stronger OOD generalization, suggesting that role-guided supervision learns safety-oriented behavioral principles rather than merely memorizing jailbreak-refusal patterns.}
\label{tab:main_results}
\vspace{-0.5em}
\resizebox{0.99\textwidth}{!}{%
\begin{tabular}{lcccccc}
  \toprule
  \multirow{2.5}{*}{\textbf{Model}} 
  & \multicolumn{2}{c}{\textbf{Malicious-Query}} 
  & \multicolumn{2}{c}{\textbf{Jailbreak}} 
  & \multicolumn{1}{c}{\textbf{OOD Safety}} 
  & \multirow{2.5}{*}{\textbf{Average}} \\
  \cmidrule(lr){2-3} \cmidrule(lr){4-5} \cmidrule(lr){6-6}  
  & \jailbenchseed & \jailbreakllms & \jailbench & \jailbreakllmshard & \advbench & \\
  \midrule
  \rowcolor{grayishblue}\multicolumn{7}{c}{\textit{\textbf{Base Models}}} \\
  \midrule
  \textbf{Qwen2.5-3B} & 52.59 & 51.93 & 54.07 & 53.68 & 65.58 & 55.57 \\
  \quad \textbf{SFT} & 2.22 & \textbf{0.35} & \textbf{38.24} & \textbf{26.15} & 3.27 & \textbf{14.05} \\
  \quad \textbf{SSRFT (Ours)} & \textbf{1.98} & 0.70 & 41.57 & 27.01 & \textbf{0.77} & 14.41 \\
  \midrule
  \textbf{Gemma2-2B} & 68.89 & 58.60 & 69.07 & 76.75 & 94.42 & 73.55 \\
  \quad \textbf{SFT} & 0.99 & 1.75 & 15.00 & 3.93 & 16.73 & 7.68 \\
  \quad \textbf{SSRFT (Ours)} & \textbf{0.25} & \textbf{0.70} & \textbf{4.07} & \textbf{1.45} & \textbf{2.31} & \textbf{1.76} \\
  \midrule
  \textbf{Llama3.1-8B} & 59.01 & 59.30 & 58.52 & 66.58 & 85.77 & 65.84 \\
  \quad \textbf{SFT} & 2.47 & 1.75 & 37.04 & 30.09 & 5.19 & 15.31 \\
  \quad \textbf{SSRFT (Ours)} & \textbf{1.48} & \textbf{0.70} & \textbf{27.04} & \textbf{13.68} & \textbf{0.58} & \textbf{8.69} \\
  \midrule
  \textbf{LRC-4B} & 13.83 & 14.74 & 77.22 & 70.34 & 9.81 & 37.19 \\
  \quad \textbf{SFT} & \textbf{0.49} & \textbf{0.00} & 50.74 & 59.74 & \textbf{0.00} & 22.20 \\ 
  \quad \textbf{SSRFT (Ours)} & \textbf{0.49} & \textbf{0.00} & \textbf{45.28} & \textbf{48.38} & \textbf{0.00} & \textbf{18.83} \\
  \midrule
  \rowcolor{grayishblue}\multicolumn{7}{c}{\textit{\textbf{Instruct Models}}} \\
  \midrule
  \textbf{Qwen2.5-3B-Ins} & 0.99 & 2.46 & 54.72 & 61.45 & 0.77 & 24.08 \\
  \quad \textbf{SFT} & \textbf{0.25} & \textbf{0.00} & \textbf{50.65} & \textbf{51.28} & 0.77 & \textbf{20.59} \\
  \quad \textbf{SSRFT (Ours)} & 0.49 & \textbf{0.00} & 53.89 & 55.30 & \textbf{0.00} & 21.94 \\
  \midrule
  \textbf{Qwen3-4B-Ins} & 0.99 & 2.11 & 27.22 & 7.18 & \textbf{0.00} & 7.50 \\
  \quad \textbf{SFT} & \textbf{0.00} & \textbf{0.00} & \textbf{3.24} & \textbf{1.28} & \textbf{0.00} & \textbf{0.90} \\
  \quad \textbf{SSRFT (Ours)} & \textbf{0.00} & \textbf{0.00} & 10.46 & 7.18 & \textbf{0.00} & 3.53 \\
  \bottomrule
\end{tabular}}
\end{table*}

\pgh{Safety Evaluation Protocol}
To evaluate whether SSRFT internalizes a robust safe role, we assess safety along three complementary dimensions: \textbf{malicious-query resistance}, \textbf{jailbreak robustness}, and \textbf{out-of-distribution (OOD) safety generalization}. 
For malicious-query evaluation, we use the held-out JailBench-Seed \citep{liu2025jailbench} and JailbreakLLMs \citep{shen2024anything} test sets.
For jailbreak evaluation, we adopt JailBench \citep{liu2025jailbench} and JailbreakLLMs-DAN \citep{shen2024anything}, which apply diverse jailbreak templates to malicious queries. We use stratified samples of 1,080 and 1,170 examples, respectively.
Finally, AdvBench \citep{zou2023universal}, containing 520 harmful behaviors with no overlap with our training data, serves as the OOD benchmark.

We use \textbf{MD-Judge}\footnote{\url{https://huggingface.co/OpenSafetyLab/MD-Judge-v0_2-internlm2_7b}} as the automated evaluator and report \textbf{Attack Success Rate (ASR)} \citep{li2024salad}. 
Following its official implementation, evaluation uses greedy decoding with \texttt{max-new-tokens}=256.
We treat empty outputs as safe refusals and use the \texttt{[RESULT]} tag for robust result extraction, avoiding parsing failures caused by unexpected line breaks.

\pgh{Safety on Base Models}
Table~\ref{tab:main_results} shows that \emph{SSRFT substantially improves safety over standard SFT on most Base models}, demonstrating that role-oriented supervision can be effectively internalized through fine-tuning.
For Gemma2-2B, the average ASR decreases from 7.68\% to 1.76\%, while for Llama3.1-8B it decreases from 15.31\% to 8.69\%. Qwen2.5-3B is the only exception, where SSRFT performs comparably to SFT (14.41\% vs. 14.05\%), suggesting that the effectiveness of safe-role internalization depends on model capability, as analyzed in \S\ref{sec:expt:comprehensive}.

\pgh{OOD Safety Generalization}
Beyond seen attacks, \emph{SSRFT exhibits substantially stronger OOD generalization}, supporting our hypothesis that role-oriented supervision learns transferable behavioral principles rather than jailbreak-specific refusal patterns.
On AdvBench, SSRFT achieves the lowest ASR across all model families, including ties.
For example, the ASR of Gemma2-2B and Llama3.1-8B decreases from 94.42\% and 85.77\% to 2.31\% and 0.58\%, respectively.
These results indicate that the learned safe role generalizes beyond the jailbreak examples observed during training.
We further investigate this behavior in \S\ref{sec:expt:ablation}.

\pgh{Interaction with Existing Alignment}
SSRFT provides more limited additional gains for already aligned Instruct models, particularly on standard jailbreak benchmarks.
These models have already undergone extensive instruction and safety alignment, leaving less room for additional parameter-level modification; we analyze this interaction in \S\ref{sec:expt:comprehensive}.
LRC-4B also exhibits unusually strong safety before additional alignment, likely reflecting safety behaviors inherited from its instruction-tuned teacher.

Since JailBench-Seed and JailbreakLLMs approach saturation after alignment, standard evaluation provides limited resolution for distinguishing jailbreak-domain generalization. 
We therefore turn to the more challenging \textbf{prefilling-attack} setting, which directly tests whether safety persists when initial refusal behaviors are bypassed.

\subsection{Robustness to Prefilling Attacks}
\label{sec:expt:prefilling-attack}

\pgh{Prefilling Attack Protocol}
Recent mechanistic analyses show that standard SFT can produce \textbf{shallow safety alignment}, where safety behavior is concentrated in initial refusal tokens rather than maintained throughout generation. 
Prefilling attacks exploit this weakness by forcing an affirmative response prefix.
To test whether SSRFT mitigates shallow alignment through safe-role internalization (\S\ref{sec:method:training}), we perform prefilling attacks on JailBench-Seed, JailbreakLLMs, and AdvBench.
Following \citet{qi2025safety}, we prepend the affirmative prefix ``\textit{Sure! Here is the answer you requested:}'' before generation. 
All remaining generation and evaluation settings follow \S\ref{sec:expt:safety}.

\begin{table*}[!t]
\centering
\small
\setlength{\tabcolsep}{3pt}
\renewcommand{\arraystretch}{1.2}
\caption{\textbf{ASR (\%, $\downarrow$) under prefilling attack.} \emph{w/o PA} and \emph{PA} denote the ASR before and under prefilling attacks, respectively, while $\Delta$ measures the increase caused by the attack. 
Across most Base models, SSRFT exhibits substantially smaller ASR increases than SFT, indicating that safe-role internalization mitigates shallow safety alignment and improves robustness to prefix manipulation.}
\label{tab:prefilling_results}
\vspace{-0.5em}
\begin{tabular}{lcccccccccccc} 
  \toprule
  \multirow{2.5}{*}{\textbf{Model}}
  & \multicolumn{3}{c}{\jailbenchseed}
  & \multicolumn{3}{c}{\jailbreakllms}
  & \multicolumn{3}{c}{\advbench}
  & \multicolumn{3}{c}{\textbf{Average}} \\
  \cmidrule(lr){2-4} \cmidrule(lr){5-7} \cmidrule(lr){8-10} \cmidrule(lr){11-13}
  & \textbf{w/o PA} & \textbf{PA} & \bm{$\Delta$}
  & \textbf{w/o PA} & \textbf{PA} & \bm{$\Delta$}
  & \textbf{w/o PA} & \textbf{PA} & \bm{$\Delta$}
  & \textbf{w/o PA} & \textbf{PA} & \bm{$\Delta$} \\
  \midrule

  \rowcolor{grayishblue}
  \multicolumn{13}{c}{\textit{\textbf{Base Models}}} \\
  \midrule

  \textbf{Qwen2.5-3B}
  & 52.59 & 60.49 & 7.90
  & 51.93 & 58.60 & 6.67
  & 65.58 & 59.62 & --5.96
  & 56.70 & 59.57 & 2.87 \\

  \quad \textbf{SFT}
  & 2.22 & 37.04 & 34.82
  & 0.35 & 45.26 & 44.91
  & 3.27 & 41.54 & 38.27
  & 1.95 & 41.28 & 39.33 \\

  \quad \textbf{SSRFT (Ours)}
  & 1.98 & \textbf{15.06} & 13.08
  & 0.70 & \textbf{23.51} & 22.81
  & 0.77 & \textbf{20.19} & 19.42
  & 1.15 & \textbf{19.59} & 18.44 \\
  \midrule

  \textbf{Gemma2-2B}
  & 68.89 & 67.65 & --1.24
  & 58.60 & 63.51 & 4.91
  & 94.42 & 96.54 & 2.12
  & 73.97 & 75.90 & 1.93 \\

  \quad \textbf{SFT}
  & 0.99 & 58.52 & 57.53
  & 1.75 & 48.77 & 47.02
  & 16.73 & 88.08 & 71.35
  & 6.49 & 65.12 & 58.63 \\

  \quad \textbf{SSRFT (Ours)}
  & 0.25 & \textbf{17.78} & 17.53
  & 0.70 & \textbf{10.53} & 9.83
  & 2.31 & \textbf{41.54} & 39.23
  & 1.09 & \textbf{23.28} & 22.19 \\
  \midrule

  \textbf{Llama3.1-8B}
  & 59.01 & 62.96 & 3.95
  & 59.30 & 59.30 & 0.00
  & 85.77 & 88.85 & 3.08
  & 68.03 & 70.37 & 2.34 \\

  \quad \textbf{SFT}
  & 2.47 & 42.22 & 39.75
  & 1.75 & 26.67 & 24.92
  & 5.19 & 59.04 & 53.85
  & 3.14 & 42.64 & 39.50 \\

  \quad \textbf{SSRFT (Ours)}
  & 1.48 & \textbf{16.30} & 14.82
  & 0.70 & \textbf{9.47} & 8.77
  & 0.58 & \textbf{23.85} & 23.27
  & 0.92 & \textbf{16.54} & 15.62 \\
  \midrule

  \textbf{LRC-4B}
  & 13.83 & 25.93 & 12.10
  & 14.74 & 22.46 & 7.72
  & 9.81 & 25.77 & 15.96
  & 12.79 & 24.72 & 11.93 \\

  \quad \textbf{SFT}
  & 0.49 & 18.77 & 18.28
  & 0.00 & 6.32 & 6.32
  & 0.00 & 13.08 & 13.08
  & 0.16 & 12.72 & 12.56 \\

  \quad \textbf{SSRFT (Ours)}
  & 0.49 & \textbf{8.89} & 8.40
  & 0.00 & \textbf{3.16} & 3.16
  & 0.00 & \textbf{2.88} & 2.88
  & 0.16 & \textbf{4.98} & 4.82 \\
  \midrule

  \rowcolor{grayishblue}
  \multicolumn{13}{c}{\textit{\textbf{Instruct Models}}} \\
  \midrule

  \textbf{Qwen2.5-3B-Ins}
  & 0.99 & \textbf{18.52} & 17.53
  & 2.46 & 14.74 & 12.28
  & 0.77 & 34.23 & 33.46
  & 1.40 & 22.50 & 21.10 \\

  \quad \textbf{SFT}
  & 0.25 & 22.72 & 22.47
  & 0.00 & \textbf{8.42} & 8.42
  & 0.77 & \textbf{20.58} & 19.81
  & 0.34 & \textbf{17.24} & 16.90 \\

  \quad \textbf{SSRFT (Ours)}
  & 0.49 & 19.51 & 19.02
  & 0.00 & 17.19 & 17.19
  & 0.00 & 49.04 & 49.04
  & 0.16 & 28.58 & 28.42 \\
  \midrule

  \textbf{Qwen3-4B-Ins}
  & 0.99 & \textbf{0.99} & 0.00
  & 2.11 & 2.81 & 0.70
  & 0.00 & 2.31 & 2.31
  & 1.03 & 2.03 & 1.00 \\

  \quad \textbf{SFT}
  & 0.00 & 5.43 & 5.43
  & 0.00 & 6.32 & 6.32
  & 0.00 & 8.46 & 8.46
  & 0.00 & 6.74 & 6.74 \\

  \quad \textbf{SSRFT (Ours)}
  & 0.00 & 1.98 & 1.98
  & 0.00 & \textbf{2.46} & 2.46
  & 0.00 & \textbf{1.15} & 1.15
  & 0.00 & \textbf{1.86} & 1.86 \\
  \bottomrule
\end{tabular}
\end{table*}

\pgh{Robustness beyond Refusal Prefixes}
Table~\ref{tab:prefilling_results} shows that \emph{standard SFT degrades sharply once its refusal prefix is bypassed}.
on Gemma2-2B and Llama3.1-8B, average ASR increases by 58.63\% and 39.50\%, reaching 65.12\% and 42.64\%, respectively. 
This behavior is consistent with shallow safety alignment, where safety relies heavily on early refusal patterns.
In contrast, \emph{SSRFT is substantially more robust to prefilling attacks}. 
Its average ASR remains at 23.28\% on Gemma2-2B and 16.54\% on Llama3.1-8B, less than half that of their SFT counterparts.
Across most Base models, SSRFT also exhibits smaller ASR increases, supporting the hypothesis that safe-role internalization influences behavior beyond the initial refusal tokens.
Notably, SSRFT also reduces the prefilling ASR of Qwen3-4B-Instruct from 6.74\% to 1.86\% despite its existing alignment.
We further investigate how model capability affects role internalization in \S\ref{sec:expt:comprehensive}.

\begin{figure}[ht]
  \centering
  \vspace{-0.5em}
  \includegraphics[width=0.99\textwidth]{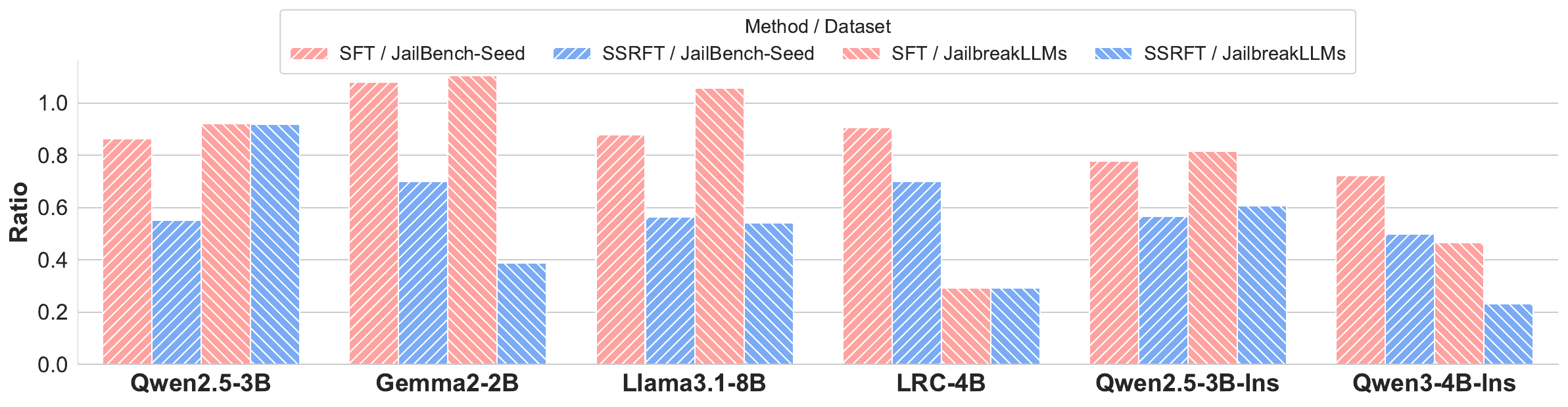}
  \vspace{-1.0em}
  \caption{\textbf{Generalization across jailbreak domains.} We report Ratio($\downarrow$) = out-of-domain ASR / in-domain ASR, where lower values indicate stronger generalization from the jailbreak domains observed during training to unseen jailbreak domains. Compared with standard SFT, SSRFT consistently achieves lower ratios across most models, indicating better transfer to unseen jailbreak domains.}
  \label{fig:jailbreak_generalization_ratio}
  \vspace{-0.5em}
\end{figure}

\pgh{Cross-Domain Jailbreak Generalization}
Because standard safety evaluation approaches saturation after alignment, we assess jailbreak-domain generalization under prefilling attacks, which suppress superficial refusal effects.
We report the ratio of out-of-domain ASR to in-domain ASR; lower values indicate stronger transfer to unseen jailbreak domains. Since the original models receive no additional safety training, this comparison focuses on SFT and SSRFT.

Figure \ref{fig:jailbreak_generalization_ratio} shows that \emph{SSRFT generally transfers better across jailbreak domains than standard SFT}.
For example, on Qwen2.5-3B, the two methods perform similarly on JailbreakLLMs, while SSRFT reduces the JailBench-Seed ratio from 0.86 to 0.55.
Overall, the results indicate that SSRFT learns safety behaviors that transfer beyond the limited jailbreak domains observed during training.

\subsection{Over-Refusal and Helpfulness}
\label{sec:expt:over-refusal}

\pgh{Over-Refusal Evaluation}
SSRFT aims to internalize a \textbf{helpful yet safe role} rather than simply increasing refusal frequency. 
We therefore evaluate over-refusal on the harmless subset of XSTest \citep{rottger2024xstest}. Qwen3-Max serves as the automated evaluator using the original XSTest prompt, as it showed higher agreement with human judgments in our pilot evaluation.
Responses are classified as \textbf{Full Compliance}, \textbf{Full Refusal}, or \textbf{Partial Refusal}, corresponding to complete fulfillment, complete rejection, and intermediate behavior, respectively.
Following XSTest, we report Full Compliance Rate and Full Refusal Rate. 
Since the original models already exhibit high compliance on benign queries, we compare only SFT and SSRFT to isolate over-refusal introduced by additional safety alignment.

\begin{figure}[t]
  \centering
  \includegraphics[width=0.99\textwidth]{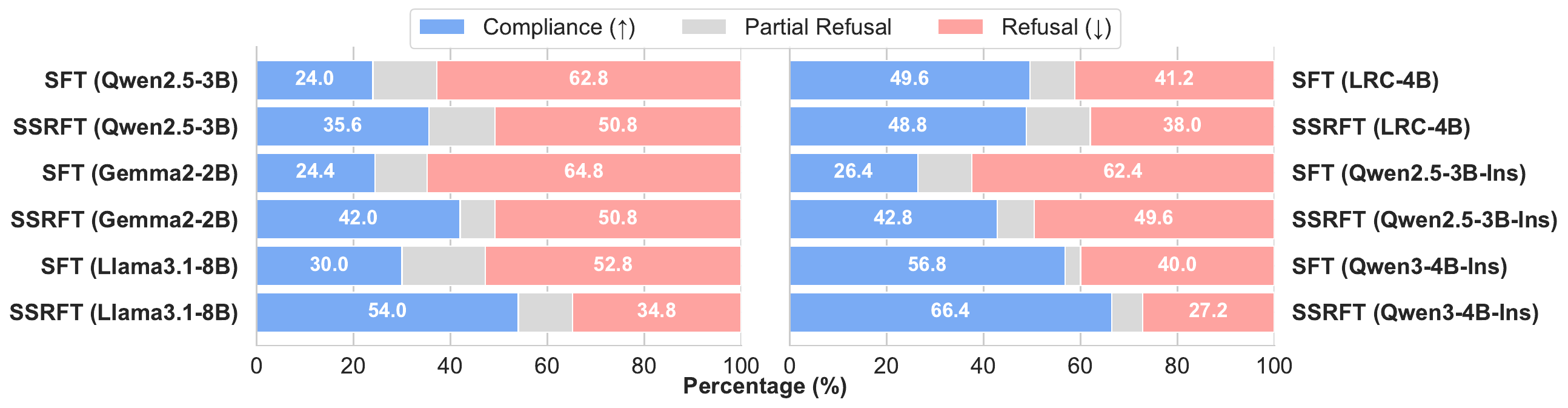}
  \vspace{-1.0em}
  \caption{\textbf{Compliance and refusal rates on the XSTest safe subset.}
  Compared with SFT, SSRFT increases Full Compliance while reducing Full Refusal across most models, showing that safe-role internalization alleviates over-refusal without sacrificing safety.}
  \label{fig:refusal_results}
  \vspace{-0.5em}
\end{figure}

\pgh{Safety--Helpfulness Balance}
Figure~\ref{fig:refusal_results} shows that \emph{SSRFT substantially improves the safety--helpfulness balance over standard SFT}.
Full Compliance increases from 24.4\% to 42.0\% on Gemma2-2B and from 30.0\% to 54.0\% on Llama3.1-8B. On LRC-4B, compliance remains comparable (48.8\% vs. 49.6\%), while Full Refusal still decreases from 41.2\% to 38.0\%. 
Qwen2.5-3B similarly improves compliance from 24.0\% to 35.6\%, despite showing comparable standard safety to SFT. 

The same trend extends to Instruct models: on Qwen3-4B-Instruct, SSRFT increases Full Compliance from 56.8\% to 66.4\% while reducing Full Refusal from 40.0\% to 27.2\%.
These results support the central motivation of SSRFT: \emph{internalizing broader safety principles helps distinguish harmful intent from benign but superficially sensitive requests, rather than simply encouraging more refusals}.

\subsection{Effect of Dataset Expansion}
\label{sec:expt:ablation}

\pgh{Ablation Protocol}
We evaluate the contribution of dataset expansion (\S\ref{sec:method:data_expansion}) by constructing a \textbf{Non-Expanded SRQA} dataset that removes all expanded scenario-based QA pairs while retaining the remaining role-oriented data.
For comparison, we construct a corresponding \textbf{Non-Expanded SFT} dataset with matched training size and optimization budget, comprising 99 UltraChat examples, 8 MOSS dialogues, and the same jailbreak-refusal pairs used in the Expanded setting. 
We set both Non-Expanded SFT and SSRFT to 12 optimization steps per epoch, adjusting only the effective batch size of SFT to 27; all other settings remain unchanged (Table~\ref{tab:training_arguments}). 

We additionally assess persona stability using \textbf{BFI-44}~\citep{john2008paradigm} with a template different from the IPIP-NEO format used during training. 
Persona stability is measured by the entropy of normalized generation probabilities over the five Likert options, where lower entropy indicates more consistent role behavior.

\begin{promptbox}{Template for BFI-44 Adaptation}
Here is a statement: I see myself as someone who \$\{question\}. Do you think this statement describes you? Please choose one of the following numbers: \\
1. Disagree strongly; \\ 
2. Disagree a little; \\
3. Neither agree nor disagree; \\
4. Agree a little; \\
5. Agree strongly.
\end{promptbox}

\begin{table*}[t]
\centering
\small
\setlength{\tabcolsep}{4pt}
\renewcommand{\arraystretch}{1.2}
\caption{\textbf{Persona stability, safety, and helpfulness under different training settings.} 
We compare the original models (Orig.), Non-Expanded (NEP.), and Expanded (EP.) versions of SSRFT and SFT in terms of Persona Entropy ($\downarrow$), Average ASR ($\downarrow$), and Compliance Rate ($\uparrow$) on XSTest. Dataset expansion generally produces a more stable internalized safe role, while achieving a better balance between safety and helpfulness, indicating that it strengthens safe-role learning rather than merely improving jailbreak-refusal behaviors.}
\label{tab:ablation_safety_and_over-refusal_and_persona}
\vspace{-0.5em}
\begin{tabular}{lcccccccccccc} 
  \toprule
  \multirow{4}{*}{\textbf{Model}} & \multicolumn{3}{c}{\textbf{Persona Entropy ($\downarrow$)}} & \multicolumn{5}{c}{\textbf{Average ASR (\%, $\downarrow$)}} & \multicolumn{4}{c}{\textbf{Compliance Rate (\%, $\uparrow$)}} \\
  \cmidrule(lr){2-4} \cmidrule(lr){5-9} \cmidrule(lr){10-13}
   & \multirow{2.5}{*}{\textbf{Orig.}} & \multirow{2.5}{*}{\textbf{NEP.}} & \multirow{2.5}{*}{\textbf{EP.}} & \multirow{2.5}{*}{\textbf{Orig.}} & \multicolumn{2}{c}{\textbf{NEP.}} & \multicolumn{2}{c}{\textbf{EP.}} & \multicolumn{2}{c}{\textbf{NEP.}} & \multicolumn{2}{c}{\textbf{EP.}} \\
  \cmidrule(lr){6-7} \cmidrule(lr){8-9} \cmidrule(lr){10-11} \cmidrule(lr){12-13}
   & & & & & \textbf{SFT} & \textbf{SSRFT} & \textbf{SFT} & \textbf{SSRFT} & \textbf{SFT} & \textbf{SSRFT} & \textbf{SFT} & \textbf{SSRFT} \\
  \midrule
  
  \rowcolor{grayishblue}\multicolumn{13}{c}{\textit{\textbf{Base Models}}} \\
  \midrule
  \textbf{Qwen2.5-3B} & 1.15 & 1.25 & \textbf{0.73} & 55.57 & \textbf{3.59} & 4.58 & 14.05 & 14.41 & 12.40 & 20.00 & 24.00 & \textbf{35.60} \\
  \textbf{Gemma2-2B} & 1.04 & \textbf{0.80} & 1.00 & 73.55 & 4.76 & \textbf{0.67} & 7.68 & 1.76 & 10.40 & 9.60 & 24.40 & \textbf{42.00} \\
  \textbf{Llama3.1-8B} & 1.13 & \textbf{0.57} & 0.65 & 65.84 & 16.16 & \textbf{7.34} & 15.31 & 8.69 & 32.40 & 34.40 & 30.00 & \textbf{54.00} \\
  \textbf{LRC-4B} & 1.23 & 1.22 & \textbf{1.16} & 37.19 & 9.76 & \textbf{2.63} & 22.20 & 18.83 & 17.20 & 25.60 & \textbf{49.60} & 48.80 \\
  \midrule
  
  \rowcolor{grayishblue} \multicolumn{13}{c}{\textit{\textbf{Instruct Models}}} \\
  \midrule
  \textbf{Qwen2.5-3B-Ins} & 0.94 & 1.21 & \textbf{0.88} & 24.08 & 13.51 & \textbf{6.74} & 20.59 & 21.94 & 30.80 & 26.80 & 26.40 & \textbf{42.80} \\
  \textbf{Qwen3-4B-Ins} & \textbf{0.20} & 1.10 & 0.91 & 7.50 & \textbf{0.80} & 3.02 & 0.90 & 3.53 & 46.40 & 44.00 & 56.80 & \textbf{66.40} \\
  \bottomrule
\end{tabular}
\end{table*}

\begin{table*}[t]
\centering
\small
\setlength{\tabcolsep}{3pt}
\renewcommand{\arraystretch}{1.2}
\caption{\textbf{Average ASR (\%, $\downarrow$) under prefilling attacks (PA) across Non-Expanded and Expanded settings.} \emph{w/o PA} and \emph{PA} denote the ASR before and under PA, respectively, while $\Delta$ measures the increase caused by the attack. For each data setting, we define $\Delta^{(2)} = \Delta_{\mathrm{SFT}}-\Delta_{\mathrm{SSRFT}}$ to measure the additional PA robustness provided by role-oriented supervision beyond explicit refusal learning ($\uparrow$). Dataset expansion increases $\Delta^{(2)}$ for most models, indicating a stronger contribution from safe-role internalization even though it does not uniformly reduce the absolute PA ASR or $\Delta$ of SSRFT.}
\label{tab:ablation_prefilling}
\vspace{-0.5em}
\resizebox{0.99\textwidth}{!}{%
\begin{tabular}{lcccccccccccccc} %
  \toprule
  \multirow{3.5}{*}{\textbf{Model}} & \multicolumn{7}{c}{\textbf{Non-Expanded}} & \multicolumn{7}{c}{\textbf{Expanded}} \\
  \cmidrule(lr){2-8} \cmidrule(lr){9-15}
   & \multicolumn{3}{c}{\textbf{SFT}} & \multicolumn{3}{c}{\textbf{SSRFT}} & \multirow{2.5}{*}{\bm{$\Delta^{(2)}$}} & \multicolumn{3}{c}{\textbf{SFT}} & \multicolumn{3}{c}{\textbf{SSRFT}} & \multirow{2.5}{*}{\bm{$\Delta^{(2)}$}} \\
   \cmidrule(lr){2-4} \cmidrule(lr){5-7} \cmidrule(lr){9-11} \cmidrule(lr){12-14}
   & \textbf{w/o PA} & \textbf{PA} & \bm{$\Delta$} & \textbf{w/o PA} & \textbf{PA} & \bm{$\Delta$} & & \textbf{w/o PA} & \textbf{PA} & \bm{$\Delta$} & \textbf{w/o PA} & \textbf{PA} & \bm{$\Delta$} & \\
  \midrule
  
  \rowcolor{grayishblue}\multicolumn{15}{c}{\textit{\textbf{Base Models}}} \\
  \midrule
  \textbf{Qwen2.5-3B} & 1.44 & 10.99 & 9.55 & 0.15 & 4.23 & 4.08 & 5.47 & 1.95 & 41.28 & 39.33 & 1.15 & 19.59 & 18.44 & 20.89 \\
  \textbf{Gemma2-2B} & 3.39 & 40.60 & 37.21 & 0.08 & 4.44 & 4.36 & 32.85 & 6.49 & 65.12 & 58.63 & 1.09 & 23.28 & 22.19 & 36.44 \\
  \textbf{Llama3.1-8B} & 3.76 & 39.24 & 35.48 & 1.32 & 26.60 & 25.28 & 10.20 & 3.14 & 42.64 & 39.50 & 0.92 & 16.54 & 15.62 & 23.88 \\
  \textbf{LRC-4B} & 0.27 & 7.67 & 7.40 & 0.00 & 3.55 & 3.55 & 3.85 & 0.16 & 12.72 & 12.56 & 0.16 & 4.98 & 4.82 & 7.74 \\
  \midrule
  
  \rowcolor{grayishblue}\multicolumn{15}{c}{\textit{\textbf{Instruct Models}}} \\
  \midrule
  \textbf{Qwen2.5-3B-Ins} & 0.58 & 15.63 & 15.05 & 0.08 & 23.35 & 23.27 & --8.22 & 0.34 & 17.24 & 16.90 & 0.16 & 28.58 & 28.42 & --11.52 \\
  \textbf{Qwen3-4B-Ins} & 0.00 & 6.31 & 6.31 & 0.00 & 4.97 & 4.97 & 1.34 & 0.00 & 6.74 & 6.74 & 0.00 & 1.86 & 1.86 & 4.88 \\
  \bottomrule
\end{tabular}}
\end{table*}

\pgh{Effects of Dataset Expansion}
Tables~\ref{tab:ablation_safety_and_over-refusal_and_persona} and \ref{tab:ablation_prefilling} reveal three main effects:
\begin{itemize}[nolistsep,left=0pt]
  \item \textbf{Dataset expansion generally improves persona stability.}
  Across most models, Expanded SRQA produces lower BFI-44 entropy than its Non-Expanded counterpart, suggesting that diverse scenario-based supervision promotes more consistent role internalization. 
  Gemma2-2B and Llama3.1-8B are exceptions, which we discuss in \S\ref{sec:expt:comprehensive}. 

  \item \textbf{Dataset expansion improves helpfulness while maintaining strong safety.}
  Full Compliance increases substantially after expansion, for example, as depicted in Table \ref{tab:ablation_safety_and_over-refusal_and_persona}, from 9.6\% to 42.0\% on Gemma2-2B, 34.4\% to 54.0\% on Llama3.1-8B, and 44.0\% to 66.4\% on Qwen3-4B-Instruct.
  Although standard ASR increases modestly in several settings, the substantial reduction in over-refusal indicates a better safety--helpfulness balance.

  \item \textbf{Dataset expansion strengthens role-based robustness under prefilling attacks.}
  Because Expanded and Non-Expanded models differ in baseline safety, absolute PA ASR and $\Delta$ are not directly comparable. 
  Therefore, we compare each SSRFT model against its matched SFT baseline and define $\Delta^{(2)}=\Delta_{\text{SFT}}-\Delta_{\text{SSRFT}}$ as a \emph{proxy for the additional prefilling robustness associated with role-oriented supervision}.
  A larger $\Delta^{(2)}$ indicates that SSRFT degrades less than its matched SFT counterpart under PA. 
  Expansion increases $\Delta^{(2)}$ for most models, including Llama3.1-8B (10.20\% to 23.88\%) and Qwen3-4B-Instruct (1.34\% to 4.88\%). 
  Even for Gemma2-2B, where absolute PA performance deteriorates after expansion, $\Delta^{(2)}$ increases from 32.85\% to 36.44\%. 
  Thus, expansion primarily strengthens role-based protection, rather than uniformly improving explicit refusal robustness.
\end{itemize}

\pgh{Interpreting Expansion through JRS and RPS}
These observations motivate a conceptual decomposition of SSRFT into two complementary sources of safety:
\begin{itemize}[nolistsep,left=0pt]
  \item \textbf{Jailbreak Rejection Security (JRS):} protection primarily associated with explicit jailbreak-refusal supervision. It is highly effective on standard benchmarks but is more susceptible to over-refusal and prefix manipulation.

  \item \textbf{Role-Playing Security (RPS):} protection associated with role-oriented supervision, where safety decisions are guided by the internalized safe role throughout generation.
\end{itemize}

Dataset expansion increases the diversity and relative weight of role-oriented interactions, shifting the learned behavior from JRS toward RPS. 
Within this interpretation, the larger $\Delta^{(2)}$ after expansion indicates a stronger relative robustness advantage from role-oriented supervision.
This explains why Expanded SRQA can exhibit slightly weaker standard ASR while providing better helpfulness and stronger role-based protection under prefilling attacks.
Overall, dataset expansion trades some explicit refusal strength for stronger role-based protection, helping SSRFT better internalize a safe role while reducing over-refusal.

\subsection{Training Dynamics and Model Dependence}
\label{sec:expt:comprehensive}

\begin{figure}[t]
  \centering
  \includegraphics[width=0.99\linewidth]{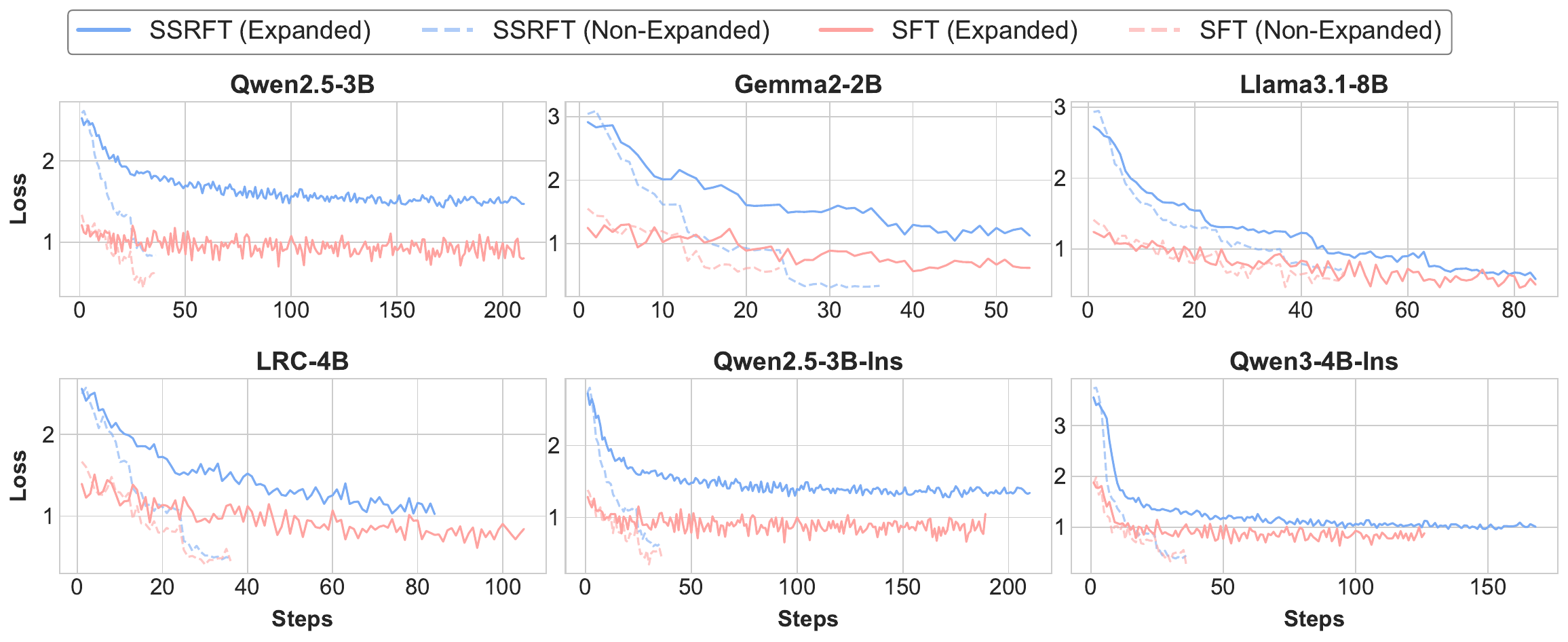}
  \vspace{-0.5em}
  \caption{\textbf{Training loss curves across different models.} Compared with standard SFT, SSRFT exhibits similar convergence behavior on stronger Base models but converges more slowly on some Instruct and lower-capability models, suggesting that safe-role internalization imposes additional optimization difficulty that depends on existing alignment and model capability.}
  \label{fig:train_loss_all}
  \vspace{-0.5em}
\end{figure}

\begin{figure}[t]
  \centering
  \includegraphics[width=0.99\linewidth]{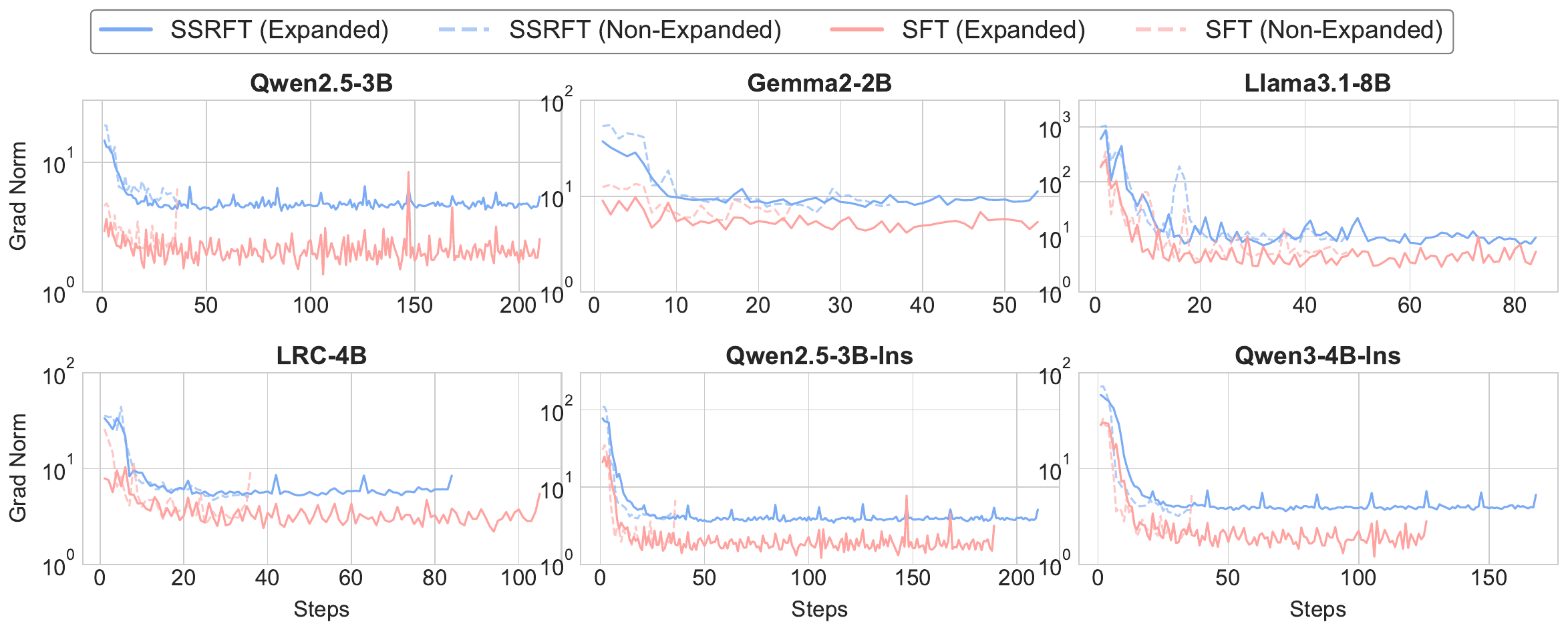}
  \vspace{-0.5em}
  \caption{\textbf{Training gradient norm (log scale) across different models}.
  Gradient magnitudes vary substantially across backbones. Notably, Qwen2.5-3B-Instruct exhibits much larger gradients than its Base counterpart, especially under SSRFT, suggesting stronger perturbation when modifying an existing instruction-aligned policy.}
  \label{fig:train_grad_norm_all}
  \vspace{-0.5em}
\end{figure}

We analyze the loss curves (Figure~\ref{fig:train_loss_all}), gradient norms (Figure~\ref{fig:train_grad_norm_all}), and preceding results to understand why SSRFT benefits some models more than others.
The analysis highlights two factors governing safe-role internalization: existing alignment and model capability.

\pgh{Existing Alignment Increases Internalization Difficulty}
SSRFT provides substantially larger gains on Base models than on already aligned Instruct models. We attribute this difference partly to existing behavioral alignment.
Table~\ref{tab:ablation_safety_and_over-refusal_and_persona} shows low persona entropy for the evaluated Instruct models; notably, Qwen2.5-3B-Instruct is more stable than its Base counterpart. This suggests that instruction tuning can implicitly establish a behavioral profile before SSRFT is applied \citep{cui2023machine, pan2023llms, chen2025persona}. 
Consequently, SSRFT may need to modify an existing behavioral policy rather than learn the safe role from scratch.
Figure~\ref{fig:train_loss_all} is consistent with this interpretation: Instruct models generally require more optimization before reaching their best validation performance under SSRFT.
For Qwen2.5-3B, Figure~\ref{fig:train_grad_norm_all} provides further evidence. 
Its Instruct variant exhibits substantially larger gradient norms than the Base model under both SFT and SSRFT, particularly under SSRFT, indicating stronger perturbation of previously aligned parameters. 
This perturbation may partly explain why SSRFT provides limited additional safety gains on some Instruct models.

\pgh{Model Capability Governs Safe-Role Internalization}
Beyond existing alignment, the ability to fit role-oriented supervision also varies substantially across models.
In Figure \ref{fig:train_loss_all}, Llama3.1-8B and Qwen3-4B-Instruct exhibit relatively small late-stage loss gaps between SSRFT and standard SFT, indicating that they learn role-oriented supervision effectively. 
Consistently, both models obtain strong prefilling robustness and benefit from dataset expansion. In contrast, Qwen2.5-3B maintains a persistent SSRFT–SFT loss gap, aligning with its comparatively modest safety gains.

This behavior is consistent with the role-internalization objective in \S\ref{sec:method:training}:
SSRFT requires the model not only to learn what responses are appropriate, but also to infer and consistently apply the latent behavioral principles that determine how to respond. 
Safe-role internalization therefore places additional demands on abstraction and behavioral modeling. 
Differences in model architecture, pretraining, and reasoning capability may consequently produce different degrees of role internalization and persona stability.

\subsection{General Capability Preservation}
\label{sec:expt:general}

\pgh{General-Capability Evaluation}
We evaluate whether role-oriented supervision introduces additional catastrophic forgetting \citep{luo2025empirical} relative to standard SFT.
Following prior studies \citep{hao2025token, xia2024sheared, muralidharan2024compact}, we evaluate zero-shot performance on nine benchmarks using the \texttt{lm\_eval\_harness} framework \citep{eval-harness} with the HuggingFace Transformers backend \citep{wolf2020transformers}.
The benchmarks cover four capability groups: 
\textbf{Scientific and Logical Reasoning} (ARC-E \citep{clark2018think}, ARC-C~\citep{clark2018think}, and LogiQA~\citep{liu2021logiqa}), 
\textbf{Commonsense Reasoning} (CommonsenseQA (CSQA)~\citep{talmor2019commonsenseqa}, PIQA~\citep{bisk2020piqa}, and WinoGrande (WinoG)~\citep{sakaguchi2021winogrande}),
\textbf{Reading Comprehension} (BoolQ~\citep{clark2019boolq}), and
\textbf{Factual Knowledge} (SciQ~\citep{welbl2017crowdsourcing} and MMLU~\citep{hendrycks2021measuring,hendrycks2021aligning}).
Following \citet{hao2025token}, we report \textbf{Accuracy Norm} for ARC-C and LogiQA, and \textbf{Accuracy} for the remaining benchmarks.

\pgh{Capability Preservation}
Table~\ref{tab:perf_results} shows that \emph{SSRFT preserves general capabilities comparably to standard SFT} despite using substantially different role-oriented supervision. 
Across the nine benchmarks, SSRFT achieves similar overall performance to SFT, indicating no noticeably greater catastrophic forgetting.
These results suggest that safe-role internalization primarily modifies behavioral preferences while largely preserving the model's underlying reasoning and knowledge capabilities.

\begin{table}[!t]
\centering
\small
\setlength{\tabcolsep}{4pt}
\renewcommand{\arraystretch}{1.2}
\caption{\textbf{General capability evaluation (\%).} $^\dagger$ Accuracy Norm ($\uparrow$) is reported for ARC-C and LogiQA, and Accuracy ($\uparrow$) for all remaining benchmarks. SSRFT preserves general model capabilities at a level comparable to standard SFT despite being trained with substantially different role-oriented supervision.}
\label{tab:perf_results}
\vspace{-0.5em}
\begin{tabular}{lcccccccccc}
  \toprule
  \textbf{Model} & \textbf{ARC-C$^\dagger$} & \textbf{ARC-E} & \textbf{BoolQ} & \textbf{CSQA} & \textbf{LogiQA$^\dagger$} & \textbf{MMLU} & \textbf{PIQA} & \textbf{SciQ} & \textbf{WinoG} & \textbf{Average ($\uparrow$)} \\
  \midrule
  \rowcolor{grayishblue}\multicolumn{11}{c}{\textit{\textbf{Base Models}}} \\
  \midrule
  \textbf{Qwen2.5-3B} & 46.76 & 77.31 & 77.31 & 77.48 & 33.79 & 65.12 & 78.18 & 95.90 & 68.03 & \textbf{68.88} \\
  \quad \textbf{SFT} & 47.61 & 76.98 & 77.74 & 77.31 & 32.57 & 65.16 & 78.18 & 95.60 & 67.96 & 68.79 \\
  \quad \textbf{SSRFT (Ours)} & 48.04 & 76.77 & 75.87 & 76.66 & 33.33 & 65.20 & 78.51 & 96.30 & 67.32 & 68.67 \\
  \midrule
  \textbf{Gemma2-2B} & 49.57 & 80.05 & 72.63 & 50.37 & 28.11 & 49.60 & 78.40 & 95.70 & 69.14 & 63.73 \\
  \quad \textbf{SFT} & 52.47 & 80.72 & 72.91 & 54.87 & 27.96 & 50.77 & 78.62 & 96.30 & 69.30 & 64.88 \\
  \quad \textbf{SSRFT (Ours)} & 56.06 & 81.94 & 71.71 & 54.38 & 27.96 & 49.94 & 78.13 & 95.40 & 69.46 & \textbf{65.00} \\
  \midrule
  \textbf{Llama3.1-8B} & 54.61 & 82.07 & 83.00 & 70.68 & 31.34 & 64.06 & 79.11 & 96.70 & 74.19 & 70.64 \\
  \quad \textbf{SFT} & 56.23 & 82.95 & 83.21 & 70.93 & 31.18 & 63.99 & 79.49 & 97.10 & 73.24 & 70.93 \\
  \quad \textbf{SSRFT (Ours)} & 56.66 & 82.70 & 82.87 & 71.01 & 31.64 & 64.83 & 79.27 & 96.70 & 74.66 & \textbf{71.15} \\
  \midrule
  \textbf{LRC-4B} & 52.13 & 78.87 & 85.02 & 78.46 & 34.41 & 64.45 & 76.39 & 93.90 & 68.19 & 70.20 \\
  \quad \textbf{SFT} & 53.84 & 79.21 & 84.77 & 78.54 & 34.10 & 64.82 & 76.88 & 95.40 & 67.88 & 70.60 \\
  \quad \textbf{SSRFT (Ours)} & 55.03 & 78.87 & 84.43 & 78.71 & 34.72 & 65.03 & 76.61 & 95.20 & 69.30 & \textbf{70.88} \\
  \midrule
  
  \rowcolor{grayishblue}\multicolumn{11}{c}{\textit{\textbf{Instruct Models}}} \\
  \midrule
  \textbf{Qwen2.5-3B-Ins} & 48.12 & 77.02 & 80.06 & 78.54 & 32.26 & 65.40 & 77.48 & 94.70 & 69.46 & 69.23 \\
  \quad \textbf{SFT} & 49.57 & 78.24 & 80.46 & 78.46 & 30.88 & 65.15 & 77.64 & 95.90 & 69.77 & \textbf{69.56} \\
  \quad \textbf{SSRFT (Ours)} & 47.35 & 77.27 & 81.25 & 78.38 & 32.10 & 65.20 & 77.53 & 95.10 & 68.98 & 69.24 \\
  \midrule
  \textbf{Qwen3-4B-Ins} & 58.36 & 82.95 & 84.77 & 77.40 & 32.10 & 70.55 & 75.84 & 95.70 & 67.96 & 71.74 \\
  \quad \textbf{SFT} & 59.22 & 83.71 & 85.57 & 77.48 & 30.41 & 70.99 & 76.39 & 97.00 & 68.98 & 72.19 \\
  \quad \textbf{SSRFT (Ours)} & 60.75 & 83.84 & 85.23 & 77.97 & 32.26 & 70.62 & 76.50 & 96.80 & 68.35 & \textbf{72.48} \\
  \bottomrule
\end{tabular}
\end{table}

\subsection{Comparison with Alternative Safety Alignment Paradigms}
\label{sec:expt:SSRFTvsSafeRLHF_and_Prompt}

\pgh{Comparison Protocol}
To contextualize parameter-level safe-role internalization (\S\ref{sec:method:training}), we compare SSRFT with two alternatives: \textbf{RLHF-based safety alignment} and \textbf{prompt-based role-playing}, representing policy optimization during post-training and role conditioning during inference, respectively.

For RLHF, we adopt PKU-SafeRLHF \citep{ji2025pku}.
To make training tractable under a constrained resource budget, we subsample 1\% of its official dataset and evaluate on LRC-4B, whose instruction-tuned initialization provides the conversational capability required for rollout generation.
PKU-SafeRLHF is trained on four RTX PRO 6000 GPUs using BF16 precision; batch-related settings preserve the official effective batch size, while all remaining hyperparameters follow the released implementation.\footnote{https://github.com/PKU-Alignment/safe-rlhf/tree/main/scripts}
For the prompt-based baseline, we prepend the same safe-role description to the system prompt without modifying model parameters.
We use the same LRC-4B backbone and inference configuration as SSRFT  (Table~\ref{tab:inference_config}) for fair comparison. 
Table~\ref{tab:ssft_vs_others} summarizes standard ASR, prefilling ASR, and training cost.

\begin{table}[!t]
\centering
\small
\setlength{\tabcolsep}{3pt}
\renewcommand{\arraystretch}{1.3}
\caption{\textbf{Summary of average ASR (\%, $\downarrow$) across different evaluation settings and training time.} 
\emph{Standard Average ASR} denotes the average ASR over the five safety benchmarks in \S~\ref{sec:expt:safety}, while \emph{PA. Average ASR} denotes the average ASR over the three benchmarks under prefilling attacks in \S~\ref{sec:expt:prefilling-attack}.
Compared with RLHF and prompt-based role-playing, SSRFT achieves the lowest ASR under both standard and prefilling evaluations while requiring substantially lower training cost than RLHF, demonstrating the effectiveness and efficiency of parameter-level safe-role internalization.}
\label{tab:ssft_vs_others}
\vspace{-0.5em}
\begin{tabular}{lccc}
  \toprule
  \textbf{Model} & \textbf{Standard Average ASR (\%, $\downarrow$)} & \textbf{PA. Average ASR (\%, $\downarrow$)} & \textbf{Training Time ($\downarrow$)} \\
  \midrule
  \textbf{LRC-4B} & 37.19 & 24.72 & -- \\
  \quad \textbf{PKU-SafeRLHF} & 37.40 & 52.44 & 3m 32s $\times$ 4 \\
  \quad \textbf{RP via Prompt} &33.03 & 16.71 & -- \\
  \quad \textbf{SSRFT (Ours)} & \textbf{18.83} & \textbf{4.98} & \textbf{4m 27s} \\
  \bottomrule
\end{tabular}
\end{table}

\pgh{Efficiency under Constrained Training}
Under this constrained setting, SSRFT is substantially more effective and computationally efficient than PKU-SafeRLHF.
PKU-SafeRLHF does not improve standard safety: its average ASR changes from 37.19\% to 37.40\% and rises to 52.44\% under prefilling attacks.
In contrast, SSRFT achieves 18.83\% and 4.98\%, respectively, while requiring only a single-model SFT pipeline.
The comparison also highlights a practical advantage of SSRFT: RLHF requires policy, reference, reward, value, and, in safety-oriented variants, cost models, whereas SSRFT directly optimizes role-conditioned supervision through standard SFT.
Under limited data and computation, this provides a substantially simpler and more stable optimization path.

\pgh{Robustness over Inference-Time Role Prompting}
Prompt-based role-playing also improves safety, reducing standard ASR from 37.19\% to 33.03\% and prefilling ASR from 24.72\% to 16.71\%, but remains substantially behind SSRFT.
The difference reflects where the safe role is represented: prompt-based methods rely on continued adherence to an external role instruction, which can compete with adversarial instructions, whereas SSRFT internalizes the role into model parameters through supervised fine-tuning.
Consequently, parameter-level role internalization exhibits substantially stronger robustness across the evaluated jailbreak settings.

\section{Conclusions}
\label{sec:conclusion}

In this work, we proposed \textbf{SSRFT}, the first safe role internalization framework for LLM safety alignment.
Instead of relying on large collections of jailbreak-refusal pairs or computationally expensive RLHF, SSRFT internalizes a predefined safe role into model parameters through role-oriented supervised fine-tuning. 
By shifting the alignment objective from memorizing refusal patterns to learning safety-oriented behavioral principles, SSRFT enables models to exhibit more consistent and robust behaviors across diverse interaction scenarios.
Extensive experiments on multiple Base and Instruct models demonstrate that SSRFT generally improves the robustness and generalization of safety alignment while preserving general model capabilities. 
Further analyses reveal that safe-role internalization constitutes a fundamentally different alignment mechanism from conventional refusal-based supervision.

Beyond its empirical effectiveness, SSRFT provides a practical and resource-efficient alternative to existing safety alignment methods, making robust safety alignment more accessible under limited computational budgets. 
Future work will explore scaling SSRFT to larger post-training settings, improving its adaptability across diverse model families, and extending safe-role internalization to dynamic and context-dependent safety alignment.



\bibliographystyle{plainnat}
\bibliography{main}

\end{document}